\documentclass[runningheads]{llncs}
\usepackage{graphicx}

\usepackage{tikz}
\usepackage{comment}
\usepackage{amsmath,amssymb} 
\usepackage{color}

\usepackage[accsupp]{axessibility}  

\usepackage{booktabs}

\usepackage{url}
\usepackage{subcaption}
\usepackage{titletoc}
\def\authcount #1 {}

\usepackage[pagebackref=true,breaklinks=true,letterpaper=true,pagebackref=true,colorlinks,linkcolor=blue,bookmarks=false]{hyperref}

\usepackage{wrapfig}

\usepackage{pifont}
\newcommand{\cmark}{\ding{51}}%
\newcommand{\xmark}{\ding{55}}%

\DeclareMathOperator*{\argmax}{arg\,max}

\newcommand{\cameraready}[1]{\textcolor{black}{#1}}

\newcommand{\topic}[1]{\vspace{1mm}\noindent\textbf{#1}}

\begin{document}
\pagestyle{headings}
\mainmatter
\def\ECCVSubNumber{5950}  

\title{Long Video Generation with Time-Agnostic VQGAN and Time-Sensitive Transformer}

\titlerunning{Long Video Generation with TATS}
%
\author{Songwei Ge\inst{1}\thanks{Work done primarily during an internship at Meta AI.} \and
Thomas Hayes\inst{2} \and
Harry Yang\inst{2} \and
Xi Yin\inst{2} \and
Guan Pang\inst{2} \\
David Jacobs\inst{1} \and
Jia-Bin Huang\inst{1,2} \and
Devi Parikh\inst{2,3}}
\authorrunning{S. Ge, T. Hayes, H. Yang, X. Yin, G. Pang, D. Jacobs, J. Huang, D. Parikh}
%
\institute{\textsuperscript{1}University of Maryland \hspace{0.2cm} \textsuperscript{2}Meta AI \hspace{0.2cm} \textsuperscript{3}Georgia Tech}
\maketitle
\begin{abstract}
Videos are created to express emotion, exchange information, and share experiences. Video synthesis has intrigued researchers for a long time. Despite the rapid progress driven by advances in visual synthesis, most existing studies focus on improving the frames' quality and the transitions between them, while little progress has been made in generating longer videos. In this paper, we present a method that builds on 3D-VQGAN and transformers to generate videos with thousands of frames. Our evaluation shows that our model trained on 16-frame video clips from standard benchmarks such as UCF-101, Sky Time-lapse, and Taichi-HD datasets can generate diverse, coherent, and high-quality long videos. We also showcase conditional extensions of our approach for generating meaningful long videos by incorporating temporal information with text and audio. Videos and code can be found at \url{https://songweige.github.io/projects/tats}.


\begin{figure*}[h]
    \centering
    \includegraphics[trim=0 0 0 0,clip,width=0.85\linewidth]{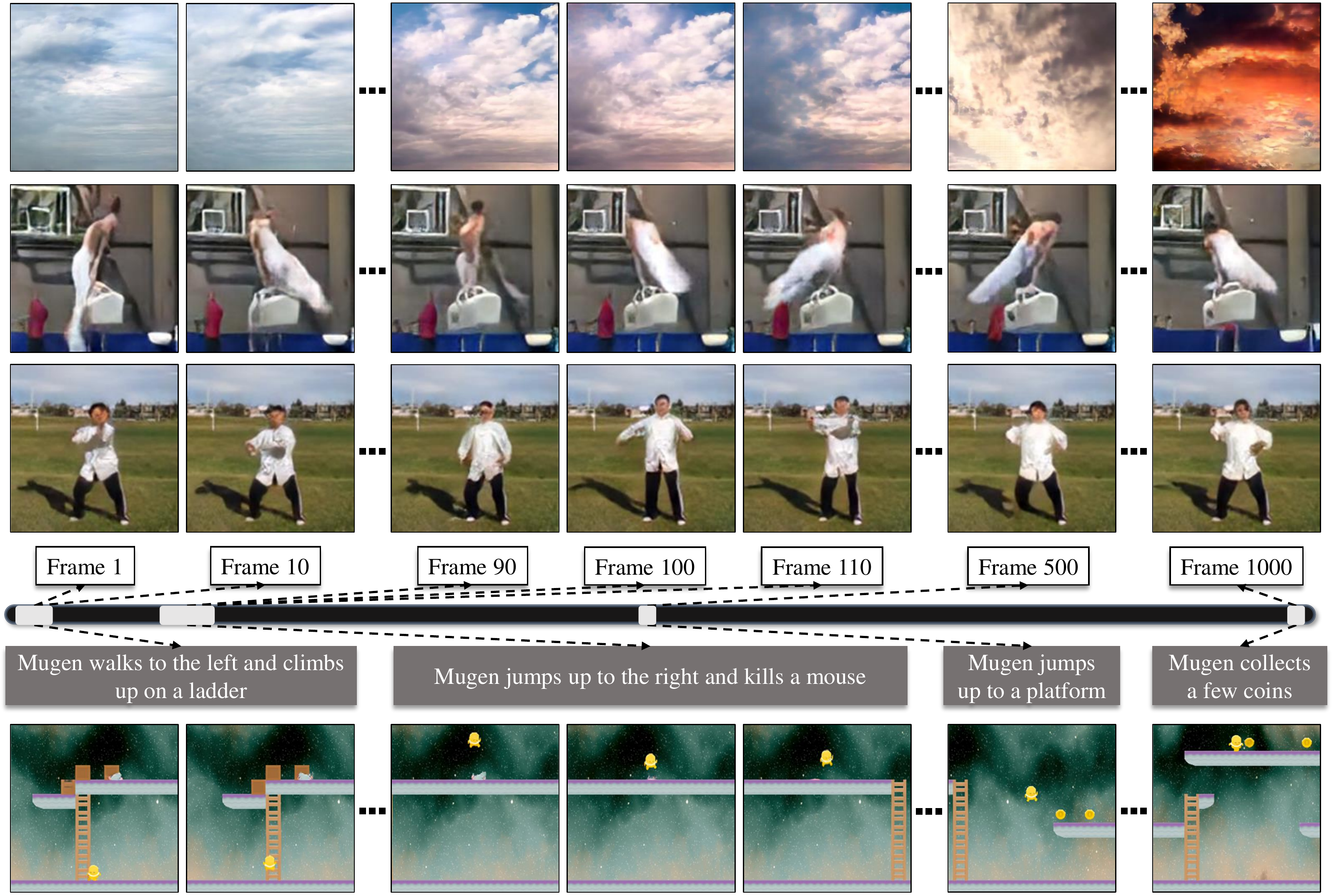}
    \caption{\textbf{Long videos generated by our model TATS with $1024$ frames.}}
    \label{fig:teaser}
\end{figure*}

\end{abstract}

\section{Introduction}

From conveying emotions that break language and cultural barriers to being the most popular medium on social platforms, videos are arguably the most informative, expressive, diverse, and entertaining visual form.
Video synthesis has been an exciting yet long-standing problem.
The challenges include not only achieving high visual quality in each frame and a natural transitions between frames, but
a consistent theme, and even a meaningful storyline throughout the video. 
The former has been the focus of many existing studies~\cite{vondrick2016generating,saito2017temporal,clark2019adversarial,tian2021a,yan2021videogpt} working with tens of frames. 
The latter is largely unexplored and requires the ability to model \emph{long-range temporal dependence} in videos with many more frames.

\begin{figure}[t]
    \centering
    \includegraphics[width=\linewidth]{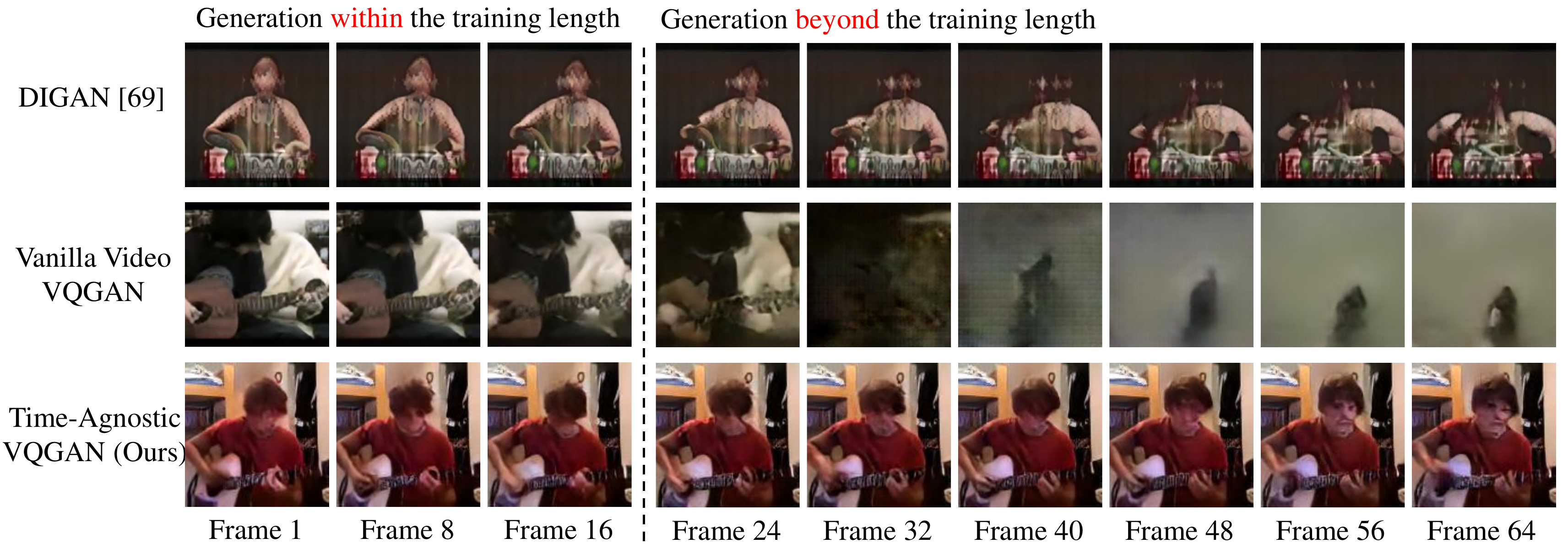}
    \caption{Video generation results with a vanilla video VQGAN and a time-agnostic VQGAN, within and beyond the training length using sliding window attention.
    }
    \label{fig:motivation}
\end{figure}

\topic{Prior works and their limitations.} 
\emph{GAN-based methods}~\cite{vondrick2016generating,saito2017temporal,clark2019adversarial,luc2020transformation} can generate plausible short videos, but extending them to longer videos requires prohibitively high memory and time cost of training and inference.\footnote{Training DVD-GAN~\cite{clark2019adversarial} or DVD-GAN-FP~\cite{luc2020transformation} on 16-frame videos requires 32-512 TPU replicas and 12-96 hours.}
\emph{Autoregressive methods} alleviate the training cost constraints through \emph{sequential prediction}. 
For example, RNNs and LSTMs generate temporal noise vectors for an image generator \cite{saito2017temporal,Tulyakov_2018_CVPR,clark2019adversarial,saito2020train,munoz2021temporal,tian2021a};
 transformers either directly generate pixel values~\cite{kalchbrenner2017video,Weissenborn2020Scaling} or indirectly predict latent tokens~\cite{van2017neural,esser2021taming,rakhimov2020latent,wu2021n,le2021ccvs,yan2021videogpt}.
These approaches circumvent training on the long videos directly by unrolling the RNN states or using a sliding window during inference. Recent works use \emph{implicit neural representations} to reduce cost by directly mapping temporal positions to either pixels~\cite{yu2021generating} or StyleGAN~\cite{karras2020analyzing} feature map values~\cite{skorokhodov2021stylegan}. 
However, the visual quality of the generated frames deteriorates quickly when performing generation beyond the training video length for all such methods as shown in Figure~\ref{fig:motivation}.

\topic{Our work.} 
We tackle the problem of long video generation. 
Building upon the recent advances of VQGAN~\cite{esser2021taming} for high-resolution \emph{image generation}, we first develop a baseline by extending the 2D-VQGAN to 3D (2D space and 1D time) for modeling videos. 
This naively extended method, however, fails to produce high-quality, coherent long videos.
Our work investigates the model design and identifies simple changes that significantly improve the capability to generate long videos of thousands of frames without quality degradation when conditioning
on no or weak information. 
Our core insights lie in 1) removing the undesired dependence on time from VQGAN 
and 2) enabling the transformer to capture long-range temporal dependence.
Below we outline these two key ideas.

\topic{Time-agnostic VQGAN.}
Our model is trained on short video clips, e.g., 16 frames, like the previous methods~\cite{vondrick2016generating,saito2017temporal,Tulyakov_2018_CVPR,clark2019adversarial,saito2020train,tian2021a}.
At inference time, we use a sliding window approach~\cite{brown2020language} on the transformer to sample tokens for a longer length. 
The sliding window repeatedly appends the most recently generated tokens to the partial sequence and drops the earliest tokens to maintain a fixed sequence length. 
However, applying a sliding attention window to 3D-VQVAE (e.g. VideoGPT~\cite{yan2021videogpt}) or 3D-VQGAN fails to preserve the video quality \emph{beyond the training length}, as shown in Figure~\ref{fig:motivation}.
The reason turns out to be that the zero paddings used in these models \emph{corrupts} the latent tokens and results in token sequences at the inference time that are drastically different from those observed during training when using sliding window.
The amount of corruption depends on the temporal position of the token.
\footnote{The large spatial span in image synthesis~\cite{esser2021taming} disguises the issue. 
When applying sliding window to border tokens, the problem resurfaces in supp. mat. Figure 12.}
We address this issue by using \emph{replicate padding} that mitigates the corruption by better approximating the real frames and brings no computational overhead.
As a result, the transformer trained on the tokens encoded by our time-agnostic VQGAN effectively preserves the visual quality \emph{beyond the training video length}.

\topic{Time-sensitive transformer.}
While removing the temporal dependence in VQGAN is desirable, long video generation certainly needs temporal information! 
This is necessary to model long-range dependence through the video and follow a sequence of events a storyline might suggest.
While transformers can generate arbitrarily long sequences, errors tend to accumulate, leading to quality degradation for long video generation. 
To mitigate this, we introduce a hierarchical architecture where an \emph{autoregressive transformer} first generates a set of sparse latent frames, providing a more global structure. Then an \emph{interpolation transformer} fills in the skipped frames autoregressively while attending to the generated sparse frames on both ends.
With these modifications, our transformer models long videos more effectively and efficiently. 
Together with our proposed 3D-VQGAN, we call our model Time-Agnostic VQGAN and Time-Sensitive Transformer (TATS). 
We highlight the capability of TATS by showing generated video samples of 1024 frames in Figure~\ref{fig:teaser}.

\topic{Our results.}
We evaluate our model on several video generation benchmarks. 
We first consider a standard short video generation setting. 
Then, we carefully analyze its effectiveness for long video generation, comparing it against several recent models. 
Given that the evaluation of long video generation has not been well studied, we generalize several popular metrics for this task considering important evaluation axes for video generation models, including long term quality and coherence.
Our model achieves state-of-the-art short and long video generation results on the UCF-101~\cite{soomro2012ucf101}, Sky Time-lapse~\cite{xiong2018learning}, Taichi-HD~\cite{siarohin2019first}, and AudioSet-Drum~\cite{gemmeke2017audio} datasets.
We further demonstrate the effectiveness of our model by conditioning on temporal information such as text and audio. 

\topic{Our contributions.} 
\begin{itemize}
    \item We identify the undesired temporal dependence introduced by the zero paddings in VQGAN as a cause of the ineffectiveness of applying a sliding window for long video generation. We propose a simple yet effective fix.
    \item We propose a hierarchical transformer that can model longer time dependence and delay the quality degradation, and show that our model can generate meaningful videos according to the story flow provided by text or audio. 
    \item To our knowledge, we are the first to generate long videos and analyze their quality. We do so by generalizing several popular metrics to a longer video span and showing that our model can generate more diverse, coherent, and higher-quality long videos.
\end{itemize}

\section{Methodology}
\label{sec:method}
\begin{figure}[t]
    \centering
    \includegraphics[width=\linewidth]{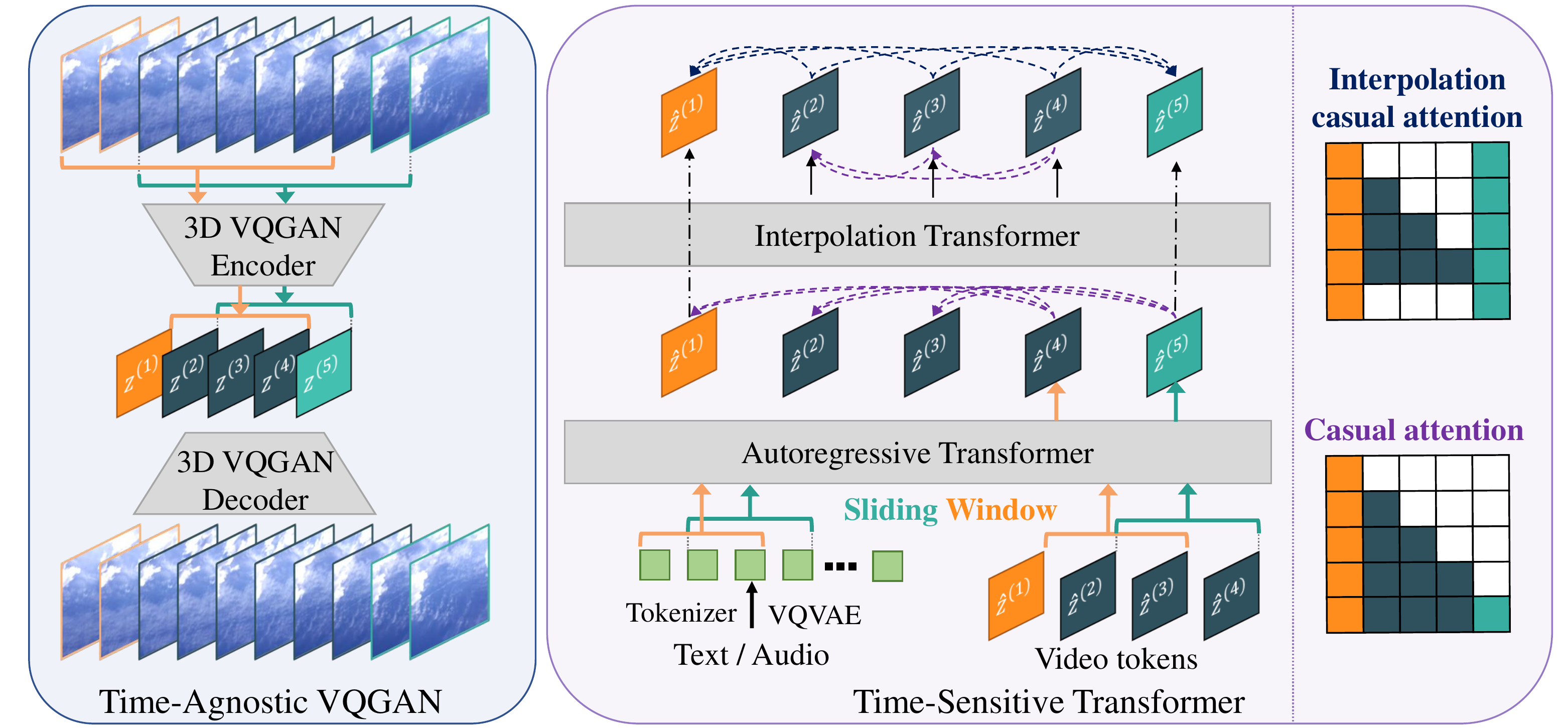}
    \caption{\textbf{Overview of the proposed framework.} Our model contains two modules: time-agnostic VQGAN and time-sensitive transformer. The former compresses the videos both temporally and spatially into discrete tokens without injecting any dependence on the relative temporal position, which allows the usage of a sliding window during inference for longer video generation. The latter uses a hierarchical transformer for capturing longer temporal dependence.}
    \label{fig:architecture}
\end{figure}

In this section, we first briefly recap the VQGAN framework and describe its extension to video generation. 
Next, we present our time-agnostic VQGAN and time-sensitive transformer models for long video generation (Figure~\ref{fig:architecture}).

\subsection{Extending the VQGAN framework for video generation}

Vector Quantised Variational AutoEncoder (VQVAE)~\cite{van2017neural}
uses a discrete bottleneck as the latent space for reconstruction. 
An autoregressive model such as a transformer is then used to model the prior distribution of the latent space. 
VQGAN~\cite{esser2021taming} is a variant of VQVAE that uses perceptual and GAN losses to achieve better reconstruction quality when increasing the bottleneck compression rates.

\topic{Vanilla video VQGAN.} 
We adapt the VQGAN architecture for video generation by replacing its 2D convolution operations with 3D convolutions. 
Given a video $\mathbf{x} \in \mathbb{R}^{T\times H\times W\times 3}$, the VQVAE consists of an encoder $f_\mathcal{E}$ and a decoder $f_\mathcal{G}$. 
The discrete latent tokens $\mathbf{z} = \mathbf{q}(f_\mathcal{E}(\mathbf{x})) \in \mathbb{Z}^{t\times h\times w}$ with embeddings $\mathbf{c}_z \in \mathbb{R}^{t\times h\times w \times c}$ are computed using a quantization operation $\mathbf{q}$ which applies nearest neighbor search using a trainable codebook $\mathcal{C}=\{\mathbf{c}_i\}_{i=1}^K$. The embeddings of the tokens are then fed into the decoder to reconstruct the input $\hat{\mathbf{x}}=f_\mathcal{G}(\mathbf{c}_z)$. 
The VQVAE is trained with the following loss:    
$$\begin{aligned}
\mathcal{L}_\text{vqvae}=\underbrace{\|\mathbf{x}-\hat{\mathbf{x}}\|_1}_{\mathcal{L}_{\text {rec}}}+\underbrace{\|\mathrm{sg}[f_\mathcal{E}(\mathbf{x})]-\mathbf{c}_z\|_{2}^{2}}_{\mathcal{L}_{\text {codebook }}}+\underbrace{\beta\|\mathrm{sg}\left[\mathbf{c}_z\right]-f_\mathcal{E}(\mathbf{x})\|_{2}^{2}}_{\mathcal{L}_{\text {commit }}}
\end{aligned},$$ 
where $\mathrm{sg}$ is a stop-gradient operation and we use $\beta=0.25$ following the VQGAN paper~\cite{esser2021taming}. 
We optimize $\mathcal{L}_{\text {codebook}}$ using an EMA update
and circumvent the non-differentiable quantization step $\mathbf{q}$ with a straight-through gradient estimator~\cite{van2017neural}. 
VQGAN additionally adopts a perceptual loss~\cite{johnson2016perceptual,zhang2018unreasonable} and a discriminator $f_\mathcal{D}$ to improve the reconstruction quality. 
Similar to other GAN-based video generation models~\cite{Tulyakov_2018_CVPR,clark2019adversarial}, we use two types of discriminators in our model - 
a spatial discriminator $f_{\mathcal{D}_s}$ that takes in random reconstructed frames $\hat{\mathbf{x}}_i \in \mathbb{R}^{H\times W\times 3}$ to encourage frame quality and a temporal discriminator $f_{\mathcal{D}_t}$ that takes in the entire reconstruction $\hat{\mathbf{x}} \in \mathbb{R}^{T\times H\times W\times 3}$ to penalize implausible motions:
$$
\mathcal{L}_\text{disc}= \log f_{\mathcal{D}_{s / t}}(\mathbf{x}) + \log(1 - f_{\mathcal{D}_{s / t}}(\hat{\mathbf{x}}))
$$
We also use feature matching losses~\cite{wang2018high,wang2018video} to stabilize the GAN training:
$$
\mathcal{L}_\text{match}=\sum_{i} p_i \left\|f_{\mathcal{D}_{s / t}/\text{VGG}}^{(i)}(\mathbf{\hat{\mathbf{x}}})-f_{\mathcal{D}_{s / t}/\text{VGG}}^{(i)}\left(\mathbf{\mathbf{x}}\right)\right\|_{1},
$$
where $f_{\mathcal{D}_{s / t}/\text{VGG}}^{(i)}$ denotes the $i^\text{th}$ layer 
of either a trained VGG network~\cite{simonyan2014very} or discriminators with a scaling factor $p_i$. 
When using a VGG network, this loss is known as the perceptual loss~\cite{zhang2018unreasonable}. 
$p_i$ is a learned constant for VGG and the reciprocal of the number of elements in the layer for discriminators.
Our overall VQGAN training objective is as follows: 
$$
\begin{aligned}
    & \min _{f_\mathcal{E},f_\mathcal{G}, \mathcal{C}}\left(\max _{f_{\mathcal{D}_s}, f_{\mathcal{D}_t}}\left(\lambda_\text{disc} \mathcal{L}_\text{disc} \right)\right)+\\
    & \min _{f_\mathcal{E},f_\mathcal{G}, \mathcal{C}}\left(\lambda_\text{match} \mathcal{L}_\text{match}+\lambda_\text{rec} \mathcal{L}_\text{rec}+ \mathcal{L}_\text{codebook}+\beta\mathcal{L}_\text{commit}\right)
\end{aligned}
$$
Directly applying GAN losses to VideoGPT~\cite{yan2021videogpt} or 3D VQGAN~\cite{esser2021taming} leads to training stability issues. In addition to the feature matching loss, we discuss other necessary architecture choices and training heuristics in supp. mat. A.1.

\topic{Autoregressive prior model.} 
After training the video VQGAN, each video can be encoded into its discrete representation $\mathbf{z}=\mathbf{q}(f_\mathcal{E}(\mathbf{x}))$. 
Following VQGAN~\cite{esser2021taming}, we unroll these tokens into a 1D sequence using the row-major 
order frame by frame. 
We then train a transformer $f_\mathcal{T}$ to model the prior categorical distribution of $\mathbf{z}$ in the dataset autoregressively:
$$p(\mathbf{z}) = p(\mathbf{z}_{0})\prod_{i=0}^{t\times h\times w-1}p(\mathbf{z}_{i+1}|\mathbf{z}_{0:i}),$$ 
where $p(\mathbf{z}_{i+1}|\mathbf{z}_{0:i}) = f_\mathcal{T}(\mathbf{z}_{0:i})$ and $\mathbf{z}_0$ is given as the start of sequence token.
We train the transformer to minimize the negative log-likelihood over training samples:
$$
\mathcal{L}_{\text {transformer }}=\mathbb{E}_{\mathbf{z} \sim p(\mathbf{z}_{\text{data}})}[-\log p(\mathbf{z})]
$$
At inference time, we randomly sample video tokens from the predicted categorical distribution $p(\mathbf{z}_{i+1}|\mathbf{z}_{0:i})$ in sequence and feed them into the decoder to generate the videos $\hat{\mathbf{x}}=f_\mathcal{G}(\hat{\mathbf{c}}_z)$. 
To synthesize videos longer than the training length, we generalize sliding attention window for our use~\cite{brown2020language} . 
A similar idea has been used in 2D to generate images of higher resolution~\cite{esser2021taming}.
We denote the $j^{th}$ temporal slice of $\mathbf{z}$ to be $\mathbf{z}^{(j)} \in \mathbb{Z}^{h\times w}$, where $0\le j \le t-1$. 
For instance, to generate $\mathbf{z}^{(t)}$ that is beyond the training length, we condition on the $t-1$ slices before it to match the transformer sequence length $p(\mathbf{z}^{(t)}|\mathbf{z}^{(1:t-1)}) = f_\mathcal{T}(\mathbf{z}^{(1:t-1)})$. 
However, as shown in Figure~\ref{fig:motivation}, when paired with sliding window attention, the vanilla video VQGAN and transformer cannot generate longer videos without quality degradation. 
Next, we discuss the reason and a simple yet effective fix.

\subsection{Time-Agnostic VQGAN}
When the Markov property holds, a transformer with a sliding window can generate arbitrarily long sequences as demonstrated in long article generation~\cite{brown2020language}. 
However, a crucial premise that has been overlooked is that the transformer needs to see sequences that start with tokens similar to $\mathbf{z}^{(1:t-1)}$ during training to predict token $\mathbf{z}^{t}$. 
This premise breaks down in VQGAN. We provide some intuitions about the reason and defer a detailed discussion to supp. mat. A.2.

Different from natural language modeling where the tokens come from realistic data, VQGAN tokens are produced by an encoder $f_\mathcal{E}$ which by default adopts zero paddings for the desired output shape.
When a short video clip is encoded, the zero paddings in the temporal dimension also get encoded and affect the output tokens, causing an unbalanced effects to tokens at different temporal position~\cite{islam2019much,kayhan2020translation,xu2021positional,alsallakh2021mind}. 
The tokens closer to the temporal boundary will be affected more significantly. 
As a result, for real data to match $\mathbf{z}^{(1:t-1)}$, they have to contain these zero-frames, which is not the case in practice. 
Therefore, removing those paddings in the temporal dimension is crucial for making the encoder \emph{time-agnostic} and enabling sliding window.

\begin{figure}[t]
    \centering
    \includegraphics[width=\linewidth]{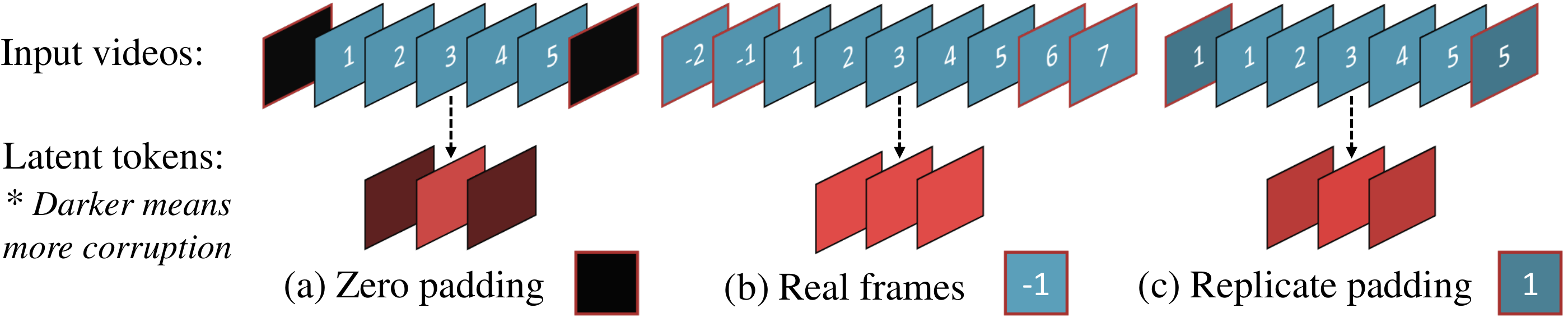}
    \caption[Caption for LOF]{\textbf{Illustration of the temporal effects induced by different paddings}. Real frame padding makes the encoder temporally shift-equivariant\protect\footnotemark ~ but introduces extra computations. Replicate padding makes decent approximation to the real frames while bringing no computational overhead.}
    \label{fig:padding_main}
\end{figure}

After removing all the padding, one needs to pad real frames to both ends of the input videos to obtain the desired output size~\cite{karras2021alias,skorokhodov2021stylegan}. 
Note that the zeros are also padded in the intermediate layers when applied, and importantly, they need not be computed. 
But, if we want to pad with realistic values, the input needs to be padded long enough to cover the entire receptive field, and all these extra values in the intermediate layers need to be computed. 
The number of needed real frames can be as large as $\mathcal{O}(Ld)$, where $L$ is the number of layers and $d$ is the compression rate in the temporal direction
of the video. 
Although padding with real frames makes the encoder fully time agnostic, it can be expensive with a large compression rate or a deep network. 
In addition, frames near both ends of the videos need to be discarded for not having enough real frames to pad, and consequently some short videos in the dataset may be completely dropped.

Instead of padding with real values, we propose to approximate real frames with the values close to them. 
An assumption, which is often referred to as the ``boring videos'' assumption in the previous literature~\cite{carreira2017quo}, is to assume that the videos are frozen beyond the given length. 
Following the assumption, the last boundary slices can be used for padding without computation. 
More importantly, this can be readily implemented by the replicate padding mode using a standard machine learning library. 
An illustration of the effects induced by different paddings can be found in Figure~\ref{fig:padding_main}. 
\footnotetext{The expressions \textit{temporally shift-equivariant} and \textit{time-agnostic} will be used interchangeably hereinafter.}
Other reasonable assumptions can also be adopted and may correspond to different padding modes. 
For instance, the reflected padding mode can be used if videos are assumed to play in reverse beyond their length, which introduces more realistic motions than the frozen frames. 
In supp. mat. A.3, we provide a careful analysis of the time dependence when applying different padding types and numbers of padded real frames. 
We find that the replicate padding alone resolves the issue well in practice and inherits the low computational cost merit of zero paddings. 
Therefore, in the following experiments, we use the replicate paddings and no real frames padded. 

\subsection{Time-Sensitive Transformer}
The time-agnostic property of the VQGAN makes it feasible to generate long videos using a transformer with sliding window attention. 
However, long video generation does need temporal information! 
To maintain a consistent theme running from the beginning of the video to the end requires the capacity to model long-range dependence. 
And besides, the spirit of a long video is in its underlying story, which requires both predicting the motions in the next few frames and the ability to plan on how the events in the video proceed. 
This section discusses the time-sensitive transformer for improved long video generation.

Due to the probabilistic nature of transformers, errors can be introduced when sampling from a categorical distribution, which then accumulate over time.
One common strategy to improve the long-term capacity is to use a hierarchical model~\cite{fan2018hierarchical,castrejon2021hierarchical} to reduce the chance of drifting away from the target theme. 
Specifically, we propose to condition one interpolation transformer on the tokens generated by the other autoregressive transformer that outputs more sparsely sampled tokens. 
The autoregressive transformer is trained in a standard way but on the sampled frames with larger intervals. 
For the interpolation transformer, it fills in the missing frames between any two adjacent frames generated by the autoregressive transformer. 
To do so, we propose interpolation attention as shown in Figure~\ref{fig:architecture}, where the predicted tokens attend to both the previously generated tokens in a causal way and the tokens from the sparsely sampled frames at both ends. 
A more detailed description can be found in supp. mat. A.4.
We consider an autoregressive transformer that generates $4\times$ more sparse video tokens. 
Generalizing this model to even more extreme cases such as video generation based on key-frames would be an interesting future work.

Another simple yet effective way to improve the long period coherence is to provide the underlying ``storyline'' of the desired video directly. 
The VQVAE framework has been shown effective in conditional image generation~\cite{esser2021taming,ramesh2021zero}. 
We hence consider several kinds of conditional information that provide additional temporal information such as audio~\cite{chatterjee2020sound2sight}, text~\cite{wu2021godiva}, and so on. 
To utilize conditional information for long video generation, we use either a tokenizer or an additional VQVAE to discretize the text or audio and prepend the obtained tokens to the video tokens and remove the start of the sequence token $\mathbf{z}_0$. 
At inference time, we extend the sliding window by simultaneously applying it to the conditioned tokens and video tokens.

\section{Experiments}
In this section, we evaluate the proposed method on several benchmark datasets for video generation with an emphasis on long video generation.
\subsection{Experimental Setups}
\topic{Datasets and evaluation.} We show results on UCF-101~\cite{soomro2012ucf101}, Sky Time-lapse~\cite{xiong2018learning}, and Taichi-HD~\cite{siarohin2019first} for unconditional or class-conditioned video generation with $128\times 128$ resolution following \cite{yu2021generating}, AudioSet-Drum~\cite{gemmeke2017audio} for audio-conditioned video generation with $64\times 64$ resolution following \cite{le2021ccvs}, and MUGEN~\cite{mugen2022mugen} with $256\times256$ resolution 
for text-conditioned video generation. We follow the previous methods~\cite{tian2021a,yu2021generating} to use Fréchet Video Distance (FVD)~\cite{unterthiner2018towards} and Kernel Video Distance (KVD)~\cite{unterthiner2018towards} as the evaluation metrics on UCF-101, Sky Time-lapse, and Taichi-HD datasets. In addition, we follow the methods evaluated on UCF-101~\cite{clark2019adversarial,le2021ccvs} to report the Inception Score (IS)~\cite{saito2020train} calculated by a trained C3D model~\cite{tran2015learning}. For audio-conditioned generation evaluation, we measure the SSIM and PSNR at the $45^\text{th}$ frame which is the longest videos considered in the previous methods~\cite{chatterjee2020sound2sight,le2021ccvs}. See 
supp. mat. B.1 for more details about the datasets.

\begin{table*}[t]
\centering\small
\caption{Quantitative results of standard video generation on different datasets. We report FVD and KVD on the Taichi-HD and Sky Time-lapse datasets, IS and FVD on the UCF-101 dataset, SSIM and PSNR at the $45^\text{th}$ frame on the AudioSet-Drum dataset.  * denotes training on the entire UCF-101 dataset instead of the train split. The class column indicates whether the class labels are used as conditional information.}
\label{tab:short-quantitative}
\begin{subtable}{.44\textwidth}
\centering\small
\caption{Sky Time-lapse} 
\begin{tabular}{lcc}
    \toprule
    Method     & FVD $(\downarrow)$  & KVD $(\downarrow)$ \\
    \midrule
    MoCoGAN-HD & 183.6\tiny{$\pm 5.2$} & 13.9\tiny{$\pm 0.7$} \\
    DIGAN  & \textbf{114.6}\tiny{$\pm 4.9$} & 6.8\tiny{$\pm 0.5$} \\ \hline
    TATS-base & 132.56\tiny{$\pm 2.6$} & \textbf{5.7}\tiny{$\pm 0.3$} \\
    \bottomrule
\end{tabular}
\caption{TaiChi-HD} 
\begin{tabular}{lcc}
    \toprule
    Method     & FVD $(\downarrow)$  & KVD $(\downarrow)$ \\
    \midrule
    MoCoGAN-HD & 144.7\tiny{$\pm 6.0$} & 25.4\tiny{$\pm 1.9$} \\
    DIGAN  & 128.1\tiny{$\pm 4.9$} & 20.6\tiny{$\pm 1.1$} \\ \hline
    TATS-base & \textbf{94.60}\tiny{$\pm 2.7$} & \textbf{9.8}\tiny{$\pm 1.0$} \\
    \bottomrule
\end{tabular}
\caption{AudioSet-Drum}
\begin{tabular}{lcc}
    \toprule
    Method & SSIM ($\uparrow$) & PSNR ($\uparrow$) \\
    \midrule
    SVG-LP & 0.510\tiny{$\pm 0.008$} & 13.5\tiny{$\pm 0.1$} \\
    \scriptsize{Vougioukas} \tiny{\textit{et al.}}  & 0.896\tiny{$\pm 0.015$} & 23.3\tiny{$\pm 0.3$} \\
    Sound2Sight & 0.947\tiny{$\pm 0.007$} & 27.0\tiny{$\pm 0.3$} \\
    CCVS & 0.945\tiny{$\pm 0.008$} & 27.3\tiny{$\pm 0.5$} \\ \hline
    TATS-base & \textbf{0.964}\tiny{$\pm 0.005$} & \textbf{27.7}\tiny{$\pm 0.4$} \\
    \bottomrule
\end{tabular}
\end{subtable}
\hfill
\begin{subtable}{.55\textwidth}
\centering\small
\caption{UCF-101}\label{tab:main_ucf} 
\begin{tabular}{lccc}
\toprule
Method & Class & IS $(\uparrow)$  & FVD  $(\downarrow)$  \\
\midrule
VGAN & \cmark & 8.31\tiny{$\pm .09$} & - \\
TGAN  & \xmark & 11.85\tiny{$\pm .07$}   & - \\
TGAN & \cmark & 15.83\tiny{$\pm .18$} & -  \\
MoCoGAN & \cmark & 12.42\tiny{$\pm .07$}   & -   \\
ProgressiveVGAN & \cmark & 14.56\tiny{$\pm .05$}   & -        \\
LDVD-GAN & \xmark & 22.91\tiny{$\pm .19$}   & - \\
VideoGPT & \xmark & 24.69\tiny{$\pm .30$}   & - \\
TGANv2 & \cmark & 28.87\tiny{$\pm .67$} & 1209\tiny{$\pm 28$}  \\
DVD-GAN*  & \cmark & 27.38\tiny{$\pm .53$} & - \\
MoCoGAN-HD*  & \xmark & 32.36 & 838 \\
DIGAN & \xmark & 29.71\tiny{$\pm .53$}   & 655\tiny{$\pm 22$}   \\
DIGAN*  & \xmark & 32.70\tiny{$\pm .35$}   & 577\tiny{$\pm 21$}   \\
CCVS*+Real frame & \xmark & 41.37\tiny{$\pm .39$} & 389\tiny{$\pm 14$} \\ 
CCVS*+StyleGAN  & \xmark & 24.47\tiny{$\pm .13$} & 386\tiny{$\pm 15$} \\
\cameraready{StyleGAN-V}*  & \xmark & 23.94\tiny{$\pm .73$} & - \\
\cameraready{CogVideo}*  & \cmark & 50.46 & 626 \\
\cameraready{Video Diffusion}*  & \xmark & 57.00\tiny{$\pm .62$} & - \\ \hline
Real data & - & 90.52 & - \\ \hline
TATS-base & \xmark & 57.63\tiny{$\pm .24$} & 420\tiny{$\pm 18$} \\
TATS-base & \cmark & \textbf{79.28}\tiny{$\pm .38$} & \textbf{332}\tiny{$\pm 18$} \\
\bottomrule
\end{tabular}
\begin{minipage}{.9\textwidth}
\end{minipage}
\end{subtable}
\end{table*}

\topic{Training details.} To compare our methods with previous ones, we train our VQGAN on videos with $16$ frames. We adopt a compression rate $d=4$ in temporal dimension and $d=8$ in spatial dimensions. For transformer models, we train a decoder-only transformer with size between GPT-1~\cite{radford2018improving} and GPT-2~\cite{radford2019language} for a consideration of computational cost. We refer to our model with a single autoregressive transformer as \textbf{TATS-base} and the proposed hierarchical transformer as \textbf{TATS-hierarchical}. For audio-conditioned model, we train another VQGAN to compress the Short-Time Fourier Transform (STFT) data into a discrete space. For text-conditioned model, we use a BPE tokenizer pretrained by CLIP~\cite{radford2021learning}. See 
supp. mat. B.2 for more details about the training and inference, \cameraready{and supp. mat. B.3. for the comparison of computational costs.}

\subsection{Quantitative Evaluation on Short Video Generation}

In this section, we demonstrate the effectiveness of our TATS-base model under a standard short video generation setting, where only 16 frames are generated for each video. The quantitative results are shown in Table~\ref{tab:short-quantitative}.

Our model achieves state-of-art FVD and KVD on the UCF-101 and Taichi-HD datasets, and state-of-art KVD on the Sky Time-lapse dataset for unconditional video generation, and improved the quality on the AudioSet-Drum for audio-conditioned video generation. \cameraready{TATS-base improves on IS by $76.2\%$ over previous method with synthetic initial frames~\cite{le2021ccvs} and is competitive against concurrent works~\cite{skorokhodov2021stylegan,hong2022cogvideo,ho2022video}.}
On the UCF-101 dataset, we also explore generation with class labels as conditional information. Following the previous method~\cite{saito2017temporal}, we sample labels from the prior distribution as the input to the generation. We find that conditioning on the labels significantly eases the transformer modeling task and boosts the generation quality, improving IS from $57.63$ to $79.28$, which is close to the upper bound shown in the ``Real data'' row~\cite{acharya2018towards,kahembwe2020lower}. This has been extensively observed in image generation~\cite{brock2018large,esser2021taming} but not quite yet revealed in the video generation.
TATS-base significantly advances the IS over the previous methods, demonstrating its power in modeling diverse video datasets.

\subsection{Quantitative Evaluation on Long Video Generation}
\begin{figure}[t]
    \centering
    \includegraphics[width=\linewidth]{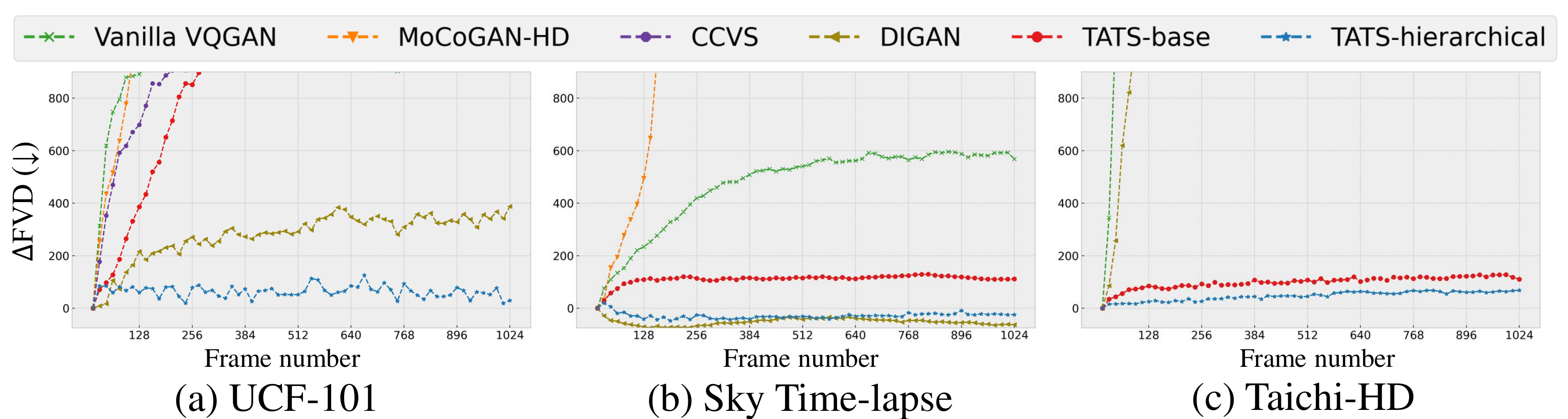}
    \caption{\textbf{Quality degradation.} FVD changes of non-overlapping 16-frame clips from the long videos generated by models trained on the UCF-101, Sky Time-lapse, and Taichi-HD datasets. The lower value indicates the slower degradation.}
    \label{fig:long_fvd_eval}
\end{figure}
Quantitatively evaluating long video generation results has been under explored. In this section, we propose several metrics by generalizing existing metrics to evaluate the crucial aspects of the long videos. We generate 512 videos with 1024 frames on each of the Sky Time-lapse, Taichi-HD, and UCF-101 datasets. We compare our proposed methods, TATS-base
and TATS-hierarchical, with the baseline Vanilla VQGAN and the state-of-the-art models including MoCoGAN-HD~\cite{tian2021a}, DIGAN~\cite{yu2021generating}, and CCVS~\cite{le2021ccvs} by unrolling RNN states, directly sampling, and sliding attention window using their official model checkpoints. 

\topic{Quality.} 
We measure video quality with respect to the duration by evaluating every 16 frames extracted side-by-side from the generated videos. Ideally, every set of 16-frame videos should come from the same distribution as the training set. 
Therefore, we report the FVD changes of these generated clips compared with the first generated 16 frames in Figure~\ref{fig:long_fvd_eval}, to measure the degradation of the video quality. The figure shows that our TATS-base model successfully delays the quality degradation compared with the vanilla VQGAN baseline, MoCoGAN-HD, and CCVS models. In addition, the TATS-hierarchical model further improves the long-term quality of the TATS-base model. The concurrent work DIGAN~\cite{yu2021generating} also claims the ability of extrapolation, while we show that the generation still degrades severely after certain number frames on the UCF-101 and Taichi-HD datasets. We conjecture the unusual decrease
of the FVD w.r.t. the duration of DIGAN and TATS-hierarchical on Sky Time-lapse can be explained by that the I3D model~\cite{carreira2017quo} used to calculate FVD is trained on Kinetics-400 dataset, and the sky videos can be outliers of the training data and lead to weak activation in the logit layers and therefore such unusual behaviors. \cameraready{We further perform qualitative and human evaluations to compare our method with DIGAN in supp. mat. C. The results confirm that the sky videos generated by TATS have better quality.}

\begin{figure}[t]
    \centering
    \includegraphics[width=\linewidth]{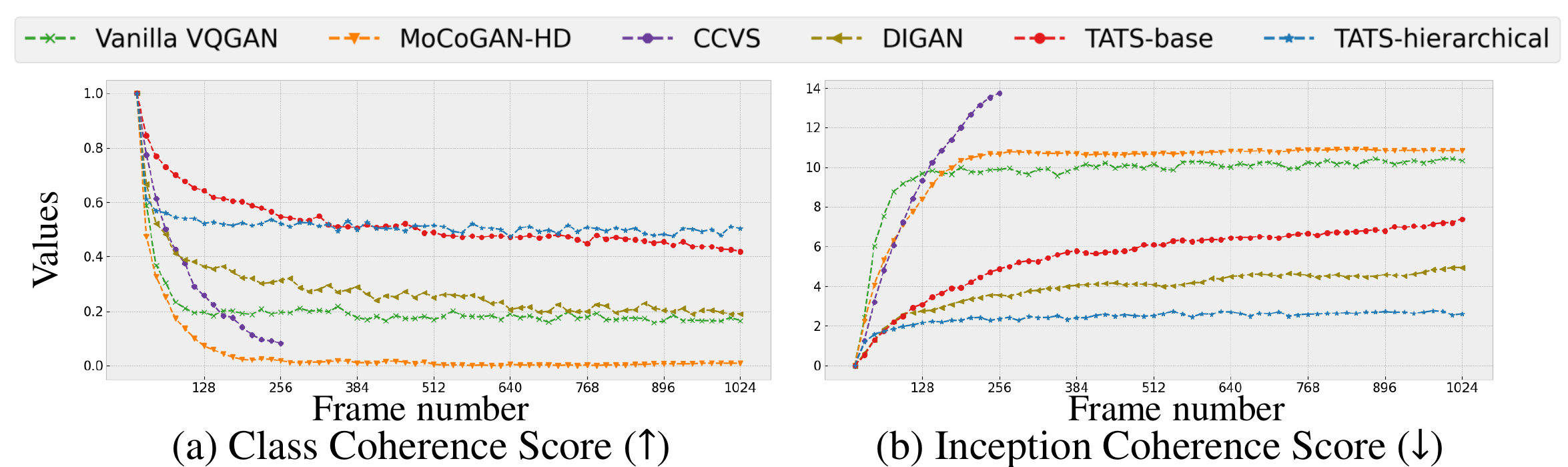}
    \caption{\textbf{Video coherency.} CCS and ICS values of every 16-frame clip extracted from long videos generated by different models trained on the UCF-101 dataset.}
    \label{fig:long_coherence_eval}
\end{figure}

\topic{Coherence.} The generated long videos should follow a consistent topical theme.
We evaluate the coherence of the videos on the UCF-101 dataset since it has multiple themes (classes). We expect the generated long videos to be classified as the same class all the time. We adopt the same trained C3D model for IS calculated~\cite{saito2020train}, and propose two metrics, Class Coherence Score (CCS) and Inception Coherence Score (ICS) at time step $t$, measuring the theme similarity between the non-overlapped 16 frames w.r.t. the first 16 frames, defined as below:
\begin{align*}
\small
\text{CCS}_t &= \sum_{i} \frac{\mathbf{1} (\argmax p_{C3D}(y|x_i^{(0:15)}), \argmax p_{C3D}(y|x_i^{(t:t+15)})}{N}
\end{align*}
\begin{align*}
\small
\text{ICS}_t &= \sum_{i} p_{C3D}(y|x_i^{(t:t+15)}) \log{\frac{p_{C3D}(y|x_i^{(t:t+15)})}{p_{C3D}(y|x_i^{(0:15)})}}
\end{align*}
The ICS captures class shift more accurately than the CCS, which only looks at the most probable class. On the other hand, CCS is more intuitive and allows us to define single metrics such as the area under the curve. CCS also doesn't have the asymmetry issue (unlike KL divergence used in ICS). We show the scores of TATS models and several baselines in Figure~\ref{fig:long_coherence_eval}. TATS-base achieves decent coherence as opposite to its quality degradation shown previously. Such difference can be explained by its failure on a small portion of generated videos as shown in supp. mat. Figure 20, which dominates the FVD. This also shows that CCS and ICS measure coherence on individual videos that complementary to FVD changes. Furthermore, TATS-hierarchical outperforms the baselines on both metrics. For example, more than half of the videos are still classified consistently in the last 16 frames of the entire 1024 frames.

\subsection{Qualitative Evaluation on Long Video Generation}

This section shows qualitative results of videos with $1024$ frames and discusses their common properties. $1024$ frames per video approaches the upper bound in the training data
as shown in supp. mat. Figure 15, 
which means the generated videos are probably different from the training videos, at least in duration.
\begin{figure}[b]
    \centering
    \includegraphics[width=\linewidth]{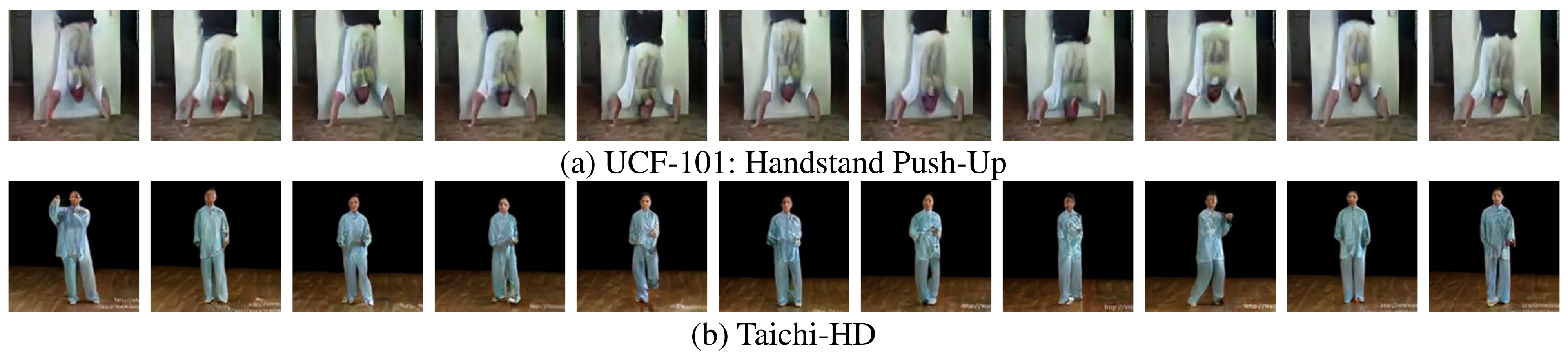}
    \caption{\textbf{Videos with recurrent events.} Every $100^\text{th}$ frame is extracted from the generated videos with $1024$ frames on the UCF-101 and Taichi-HD datasets. }
    \label{fig:self-loop}
\end{figure}
\topic{Recurrent actions.} We find that some video themes contain repeated events such as the \textit{Handstand Push-Up} class in the UCF-101 dataset and videos in the Taichi-HD dataset. As shown in Figure~\ref{fig:self-loop}, our model generalizes to long video generation by producing realistic and recurrent actions. 
However, we find that all the $1024$ generated frames in these videos are unique and motions are not identical, which shows that our model is not simply copying the short loops. 

\begin{wrapfigure}[8]{r}{0.45\textwidth}
\centering
  \footnotesize
  \begin{tabular}{lll}
    \toprule
    Videos  & LPIPS metric & Color Similarity  \\
    \midrule
    Real & $0.1839 \pm 0.0683$ & $0.7330 \pm 0.2401$ \\
    Fake & $0.3461 \pm 0.1184$ & $0.0797 \pm  0.1445$\\
     \bottomrule
    \end{tabular}
  \captionof{table}{LPIPS and color histogram correlation between the first and the last frames of the sky videos.}
  \label{tab:lpips}
\end{wrapfigure}
\begin{figure}[t]
    \centering
    \includegraphics[width=\linewidth]{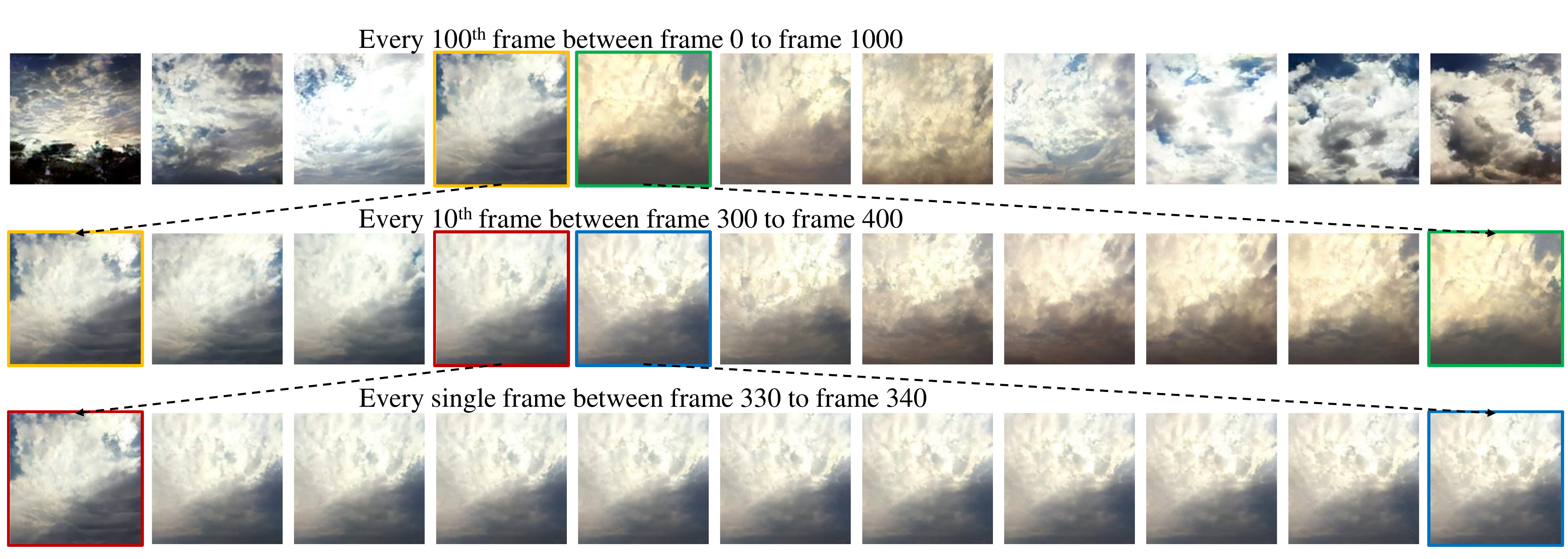}
    \caption{\textbf{Videos with homomorphisms.} Every $100^\text{th}$, $10^\text{th}$, and consecutive frames are extracted from a generated sky video with $1024$ frames.}
    \label{fig:intersection}
\end{figure}

\topic{Smooth transitions.} Videos with the same theme often share visual features such as scenes and motions. We show that with enough training data available for a single theme, our model learns to ``stitch'' the long videos through generating smooth transitions between them. For example in Figure~\ref{fig:intersection}, we show that our model generates a long sky video containing different weather and timing while the transitions between these conditions are still natural and realistic. To show that this is not the case in the training data, we compute the LPIPS score~\cite{zhang2018unreasonable} and color histrogram correlation between the $1^\text{st}$ and the $1024^\text{th}$ frames and report the mean and standard deviation on the $500$ generated and $216$ real long videos in Table~\ref{tab:lpips}. It shows that such transitions are much more prominent on the generated videos than the training videos.
Similar examples of the UCF-101 and Taichi-HD videos can be found in supp. mat. Figure 19.
A separation of content and motion latent features~\cite{Tulyakov_2018_CVPR,yu2021generating,tian2021a} may better leverage such transitions, which we leave as future work.

\begin{figure}[t]
    \centering
    \includegraphics[width=\linewidth]{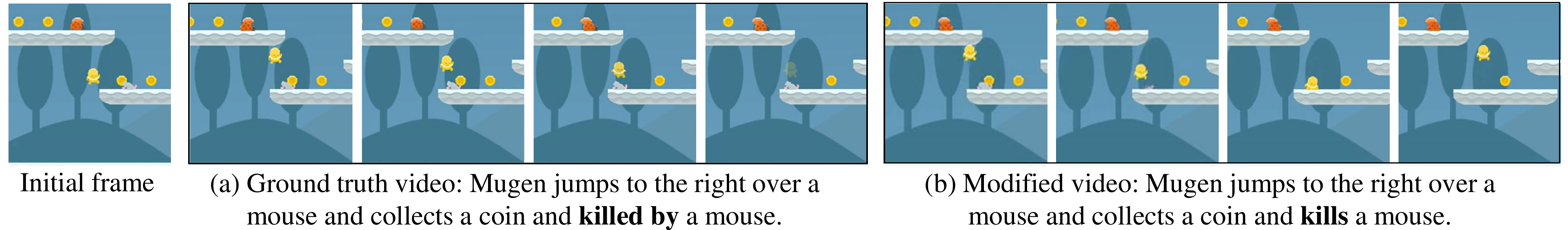}
    \caption{\textbf{Videos with meanings.} Video manipulation by modifying the texts.}
    \label{fig:manipulation}
\end{figure}

\topic{Meaningful synthesis.} By conditioning on the temporal information, we can achieve more controllable generation, which allows us to directly create or modify videos based our own will. 
For example, in Figure~\ref{fig:manipulation}, we show that it is possible to manipulate the videos by changing the underlying storyline - by replacing ``killed by'' with ``kills'' we completely change the destiny of Mugen!

\section{Related Work}



In this section, we discuss the related work in video generation using different models. 
We focus on the different strategies adopted by these methods to handle temporal dynamics and their potential for and challenges associated with long video generation. Also see supp. mat. D for other relevant models and tasks.

\topic{GAN-based video generator.} Adapting GANs~\cite{karras2019style,karras2020analyzing,goodfellow2014generative} to video synthesis requires modeling the temporal dimension. Both 3D deconvolutionals~\cite{vondrick2016generating} and additional RNN or LSTM~\cite{saito2017temporal,Tulyakov_2018_CVPR,Tulyakov_2018_CVPR,clark2019adversarial,saito2020train,munoz2021temporal,tian2021a} have been used. By unrolling the steps taken by the RNNs, videos of longer duration can be generated. However, as shown in our experiments, the quality of these videos degrades quickly. 
In addition, without further modifications, the length of videos generated by these models is limited by the GPU memory (e.g., at most 140 frames can be generated on 32GB V100 GPU by HVG~\cite{castrejon2021hierarchical}).

\topic{AR-based video generator.} Autoregressive models have become a ubiquitous generative model for video synthesis~\cite{ranzato2014video,srivastava2015unsupervised,kalchbrenner2017video,Weissenborn2020Scaling}. 
A common challenge faced by AR models is their slow inference speed. This issue is mitigated by training on the compressed tokens with VQVAEs~\cite{rakhimov2020latent,wu2021n,le2021ccvs,yan2021videogpt}. 
Our model falls into this line of video generators. We show that such a VQVAE-based AR model is promising to generate long videos with long-range dependence.

\topic{INR-based video generator.} Implicit Neural Representations~\cite{sitzmann2020implicit,tancik2020fourier} represent continuous signals such as videos by mapping the coordinate space to RGB space~\cite{yu2021generating,skorokhodov2021stylegan}. 
The advantage of these models is their ability to generate arbitrarily long videos non-autoregressively. However, their generated videos still suffer from the quality degradation or contain periodic artifacts due to the positional encoding and struggle at synthesizing new content. 

\topic{Concurrent works on long video generation.} \cameraready{In parallel to our work, \cite{ho2022video} proposes a gradient conditioning method for sampling longer videos with a diffusion model, \cite{nash2022transframer} and  \cite{hong2022cogvideo} explore a hierarchical model with frame-level VQVAE embeddings or RGB frames. \cite{brooks2022generating} uses a low-resolution long video generator and short-video super-resolution network to generate videos of dynamic scenes.} 


\section{Conclusion}
We propose TATS, a time-agnostic VQGAN and time-sensitive transformer model, that is only trained on clips with tens of frames and can generate thousands of frames using a sliding window during inference time. Our model generates meaningful videos when conditioned on text and audio. Our paper is a small step but we hope it can encourage future works on more interesting forms of video synthesis with a realistic number of frames, perhaps even movies.

\topic{Acknowledgements} We thank Oran Gafni, Sasha Sheng, and Isabelle Hu for helpful discussion and feedback; Patrick Esser and Robin Rombach for sharing additional insights for training VQGAN models; Anoop Cherian and Moitreya Chatterjee for sharing the pre-processing code for the AudioSet dataset; Jay Wu and Jaehoon Yoo for testing the FVD scores of released UCF-101 models.

\clearpage
\appendix

\startcontents[Supplementary Material]

\printcontents[Supplementary Material]{l}{1}{\section*{Supplementary Material of TATS}\setcounter{tocdepth}{2}}

\section{Implementation Details}
In this section, we provide additional details on our proposed TATS model, including 
designs to stabilize training 3D-VQGAN with GAN losses, 
discussions on the undesired temporal dependence induced by the zero padding, 
ablations on different potential solutions, 
and description of our interpolation attention.

\subsection{Training video VQGAN with GAN losses}
\label{sec:video_vqgan}
\begin{figure}[h]
    \centering
    \includegraphics[width=\linewidth]{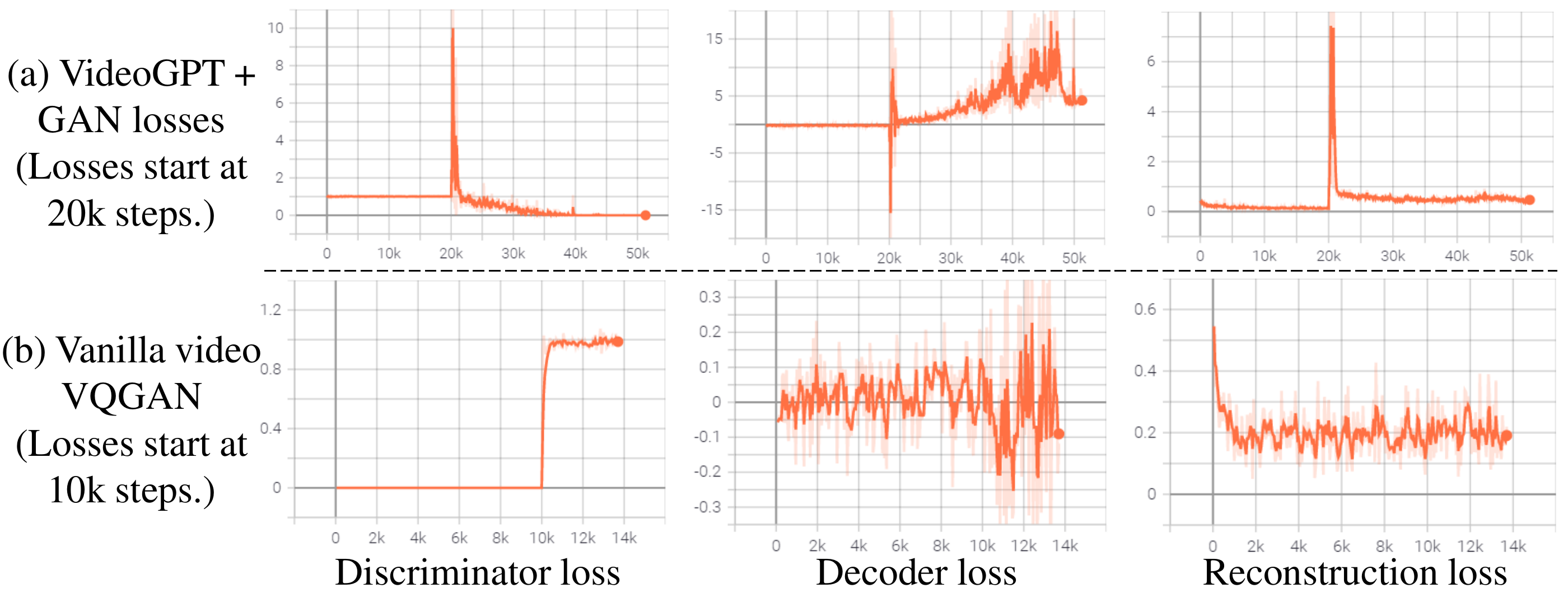}
    \caption{Training VideoGPT with GAN losses leads to discriminator collapse.}
    \label{fig:discriminator}
\end{figure}
In this section, we describe how we train 3D-VQGAN with GAN losses and clarify several major architecture choices of our proposed vanilla video VQGAN compared with VQGAN~\cite{esser2021taming} and VideoGPT~\cite{yan2021videogpt}. 
We find that directly applying GAN losses to VideoGPT leads to severe discriminator collapse. 
As shown in Figure~\ref{fig:discriminator} (a), after adding GAN losses at the 20k step, the discriminator losses saturate quickly and thus provide nonsensical gradient to the decoder. 
As a result, the decoder (generator) loss and reconstruction loss entirely fall apart. 
We found that the tricks proposed in VQGAN~\cite{esser2021taming} such as starting GAN losses at a later step and using adaptive weight, do not help the situation. 
We present several empirical methods we found effective in stabilizing the GAN losses.

\begin{figure}[h]
    \centering
    \includegraphics[width=0.95\linewidth]{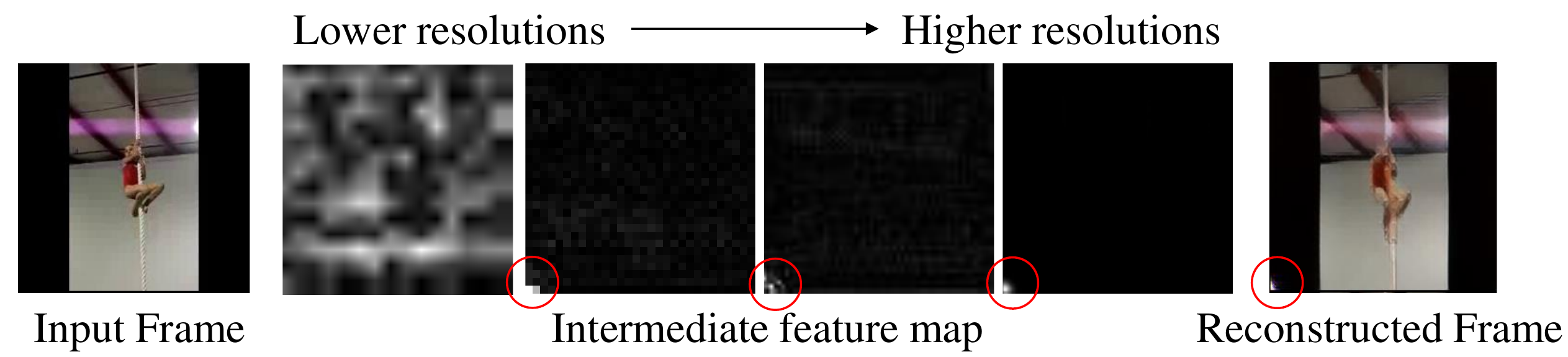}
    \caption{Blob-shaped artifacts due the the normalization layers in VQGAN.}
    \label{fig:blob}
\end{figure}

First, we find that the axial-attention layers introduced in VideoGPT interact poorly with the GAN losses and exacerbate the collapse, which is also noticed in recent work on training ViT with GAN losses~\cite{lee2021vitgan}. 
Therefore, we utilize a pure convolution architecture similar to VQGAN~\cite{esser2021taming} for video compression. 
Second, a more powerful decoder helps the reconstruction follow the discriminator closely.
We doubled the number of feature maps whenever the resolutions are halved, in contrast to VideoGPT where all the layers have a constant number of channels. 
As a consequence, similar to the previous observation~\cite{child2020very}, we find that training large VAE models would cause exploded gradients. 
Furthermore, similar to the proposed solution of gradient skipping~\cite{child2020very}, we always clip the gradient to have the Euclidean norm equal to $1$ during the training. 
Last, we notice that the blob-shaped artifacts often appear in the reconstruction and exaggerate along with the intermediate feature maps as shown in Figure~\ref{fig:blob}, which was also observed in StyleGANs~\cite{karras2019style,karras2020analyzing} and can be attributed to the normalization layers.
This is especially pronounced in the training video VQGAN due to the small batch sizes. 
We use Synced Batch Normalization as a replace of Group Normalization~\cite{wu2018group} used in original VQGAN~\cite{esser2021taming} and accumulate gradients across multiple steps and successfully mitigate this issue. 
Our vanilla video VQGAN can be steadily trained with the GAN losses with the above choices of architecture designs.

\subsection{Zero paddings inhibit sliding attention window}
\label{sec:time_agnostic_detail}

\begin{figure}[h]
    \centering
    \includegraphics[width=\linewidth]{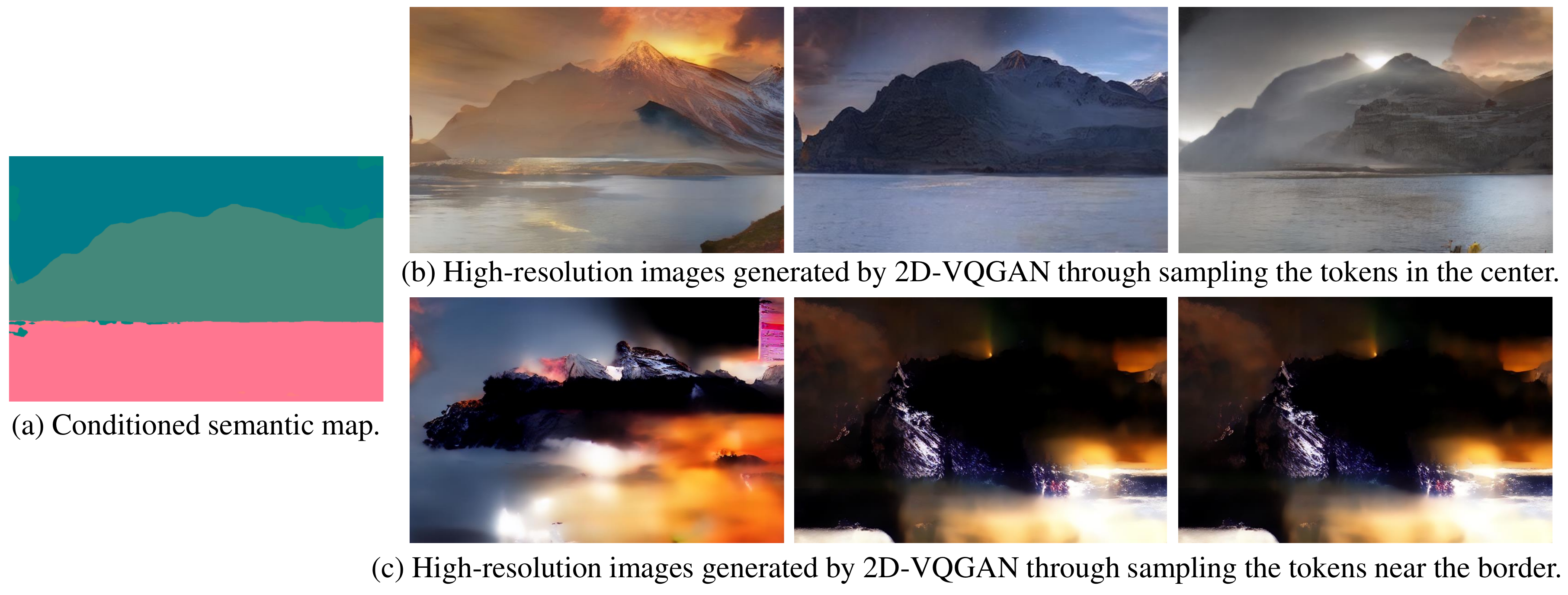}
    \caption{2D-VQGAN generates larger images by generating tokens in the center.}
    \label{fig:2dvqgan}
\end{figure}

Following the intuition in method section, we provide a more detailed discussion on why zero paddings in VQGAN inhibit the usage of the sliding attention window. 
We aim to show that when zero padding is used, there are barely tokens starting with $\mathbf{z}^{(1:t-1)}$ in the training set. 
However, this is necessary for the transformer to generate $\mathbf{z}^{(t)}$ using a sliding window. 

For simplicity, we absorb the quantization step $\mathbf{q}$ into $f_\mathcal{E}(\mathbf{x})$. 
In order for there to be tokens starting with $\mathbf{z}^{(1:t-1)}$ in the transformer training data, there need to be real video clips that can be encoded in $\mathbf{z}^{(1:t)}$. 
It is desired that $f_\mathcal{E}$ is temporally shift-equivariant given 3D convolutions so that $\mathbf{z}^{(1:t)}$ is the output of the clips slightly shifted from the original position, i.e. $\mathbf{z}^{(1:t)} = f_\mathcal{E}(\mathbf{x}^{(d:T+d-1)})$, where $d$ is the compression rate of $f_\mathcal{E}$ in the temporal dimension. 
However, we find that the encoder is \emph{not} temporally shift-equivariant and encodes $\mathbf{x}^{(d:T-1)}$ differently when these frames are positioned at different places, i.e.
\begin{equation}
    \label{eq:full_consistency}
    [f_\mathcal{E}(\mathbf{x}^{(0:T-1)})]^{(1:t-1)} \neq [f_\mathcal{E}(\mathbf{x}^{(d:T+d-1)})]^{(0:t-2)}
\end{equation} 
To see this theoretically, we consider a shift-equivariant version of $f_\mathcal{E}$ by moving all the internal zero paddings to the input. 
We denote this encoder as $\hat{f}_\mathcal{E}$ such that
\begin{equation}
    \label{eq:equal_encoder}
    f_\mathcal{E}(\mathbf{x}) = \hat{f}_\mathcal{E}([\mathbf{0}^N \; \mathbf{x} \; \mathbf{0}^N]),
\end{equation}
where $[\mathbf{0}^N \; \mathbf{x} \; \mathbf{0}^N]$ is the concatenation of $\mathbf{x}$ with $N$ zero paddings $\mathbf{0} \in \mathbb{R}^{h\times w}$. 
One can show that the number of paddings needed is $N = \mathcal{O}(Ld)$, 
where $L$ is the number of convolutional layers in the encoder. 
Given the shift equivariance of $\hat{f}_\mathcal{E}$, we can derive the desired latent tokens for transformer training as 
$$z^{(1:t)} = \hat{f}_\mathcal{E}([\mathbf{0}^{N-d} \; \mathbf{x}^{(d:T-1)} \; \mathbf{0}^{N+d}]).$$ 
However, we cannot find such real videos corresponding to $\hat{\mathbf{x}} = [\mathbf{0}^{N-d} \; \mathbf{x}^{(d:T-1)} \; \mathbf{0}^{N+d}]$ as the input to $f_\mathcal{E}$ based on the Equation~\ref{eq:equal_encoder} for two reasons. 
First, according to Equation~\ref{eq:equal_encoder} the input to $f_\mathcal{E}$ should be $\hat{\mathbf{x}}^{(N:N+T)}=[\mathbf{x}^{(T+d:T)} \; \mathbf{0}^{d}]$. 
There are rare videos whose last $d$ frames are blank in the real datasets. 
In addition, $\hat{\mathbf{x}}^{(N-d:N)}=\mathbf{x}^{(d:T+d)}$ indicates that real frames are used to pad the input, while $f_\mathcal{E}$ only uses zeros. 
Therefore, we show that no such tokens are starting with $\mathbf{z}^{(1:t-1)}$ in the training set. As a result, the transformer cannot generalize to the sequence starting with $z^{(1:t-1)}$ using a sliding attention window.

This problem occurs in all the generative models that utilize VQVAE and transformers. 
However, it is often disguised and ignored in previous studies. 
In practice, the severity of this issue depends on the receptive field, and the relative position of the generated token to the border. 
In the case of the original VQGAN~\cite{esser2021taming} that uses a sliding window to generate high-resolution images, this issue is disguised as the spatially centered token is always chosen to generate, which is far away from the border and much less affected by the zero-padding given the large spatial size (256).
We show in Figure~\ref{fig:2dvqgan} that, when generating the tokens near the border using a sliding window using the high-resolution VQGAN, the quality of images degrades quickly, similar to our observation in video generation. 
Furthermore, tokens at any position are close to the borders for synthesizing long videos due to the small temporal length (16).

\subsection{Quantifying the time agnostics of different padding types}
\label{sec:padding}

\begin{figure}[h]
    \centering
    \includegraphics[width=\linewidth]{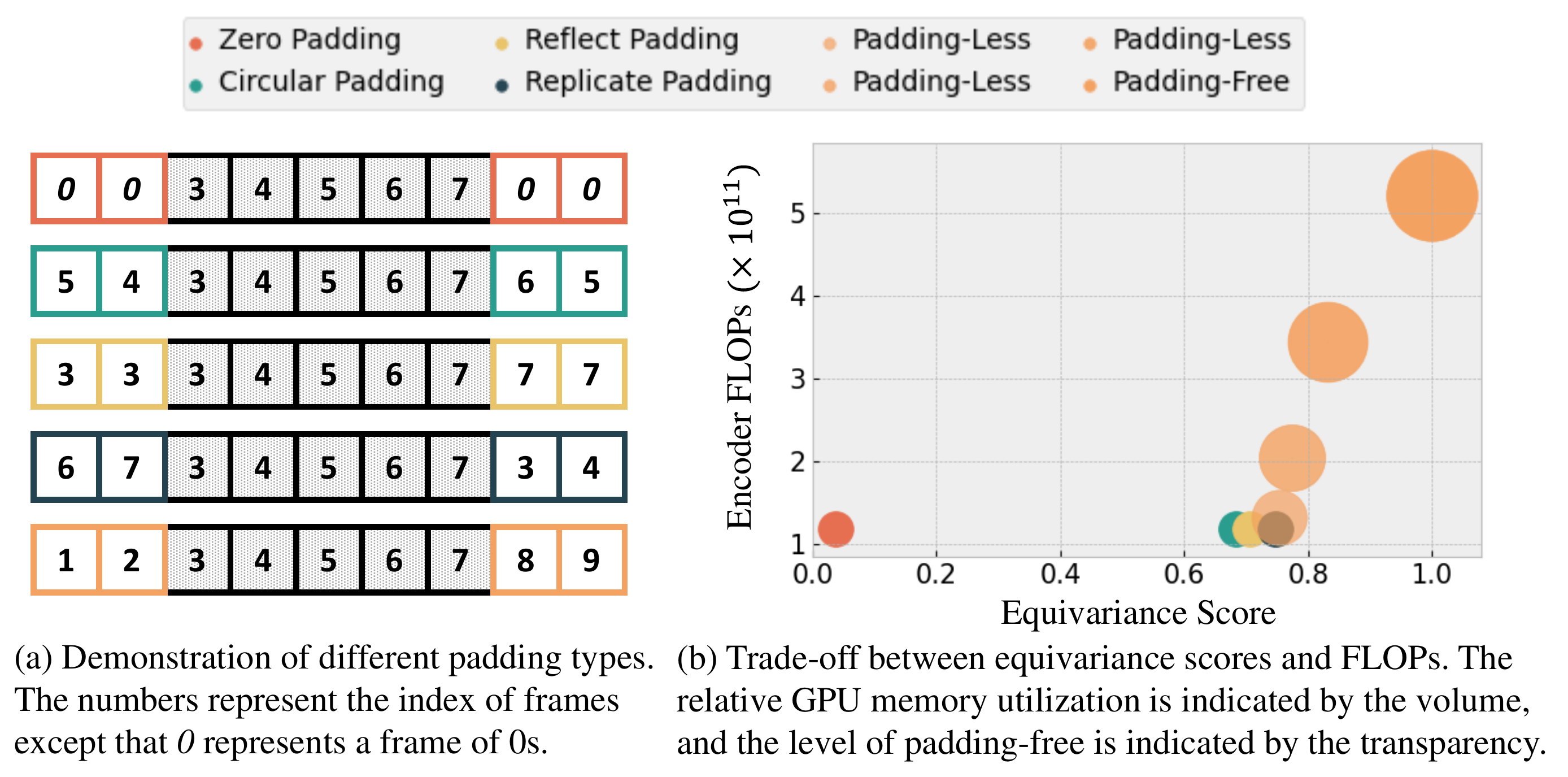}
    \caption{Demonstration of different padding types and their computational costs as well as effects on the consistency score. Note that when using less or no paddings, extra real paddings are added to the input videos.}
    \label{fig:tradeoff}
\end{figure}

Adding real frames makes the VQGAN fully time-agnostic but increases the computational cost. 
Using other paddings is less effective but brings no overheads. 
Therefore, it is essential to quantify the temporal dependence to understand the trade-off between the desired property and the cost to achieve it. 
To that end, we propose an equivariance score as a measure of time agnostic shown in Equation~\ref{eq:full_consistency}, which calculates the percentage of tokens that are identical when the same frames are positioned at the beginning of the end of the clips, which are the two extreme cases that are mostly affected by the paddings from either side:
\begin{equation}
    \sum_{i=1}^{t} \sum_{j=0}^{h-1} \sum_{k=0}^{w-1} \frac{\mathbf{1}\left([f_\mathcal{E}(\mathbf{x}_{0:T})]_{i, j, k}, [f_\mathcal{E}(\mathbf{x}_{d:T+d})]_{i-1, j, k}\right)}{t\times h\times w},
\end{equation}
where $\mathbf{1}$ is an identity function.
We report the mean and standard deviation across $1024$ clips 
using a video VQGAN trained on the UCF-101 dataset across different padding strategies. 
We visualize the trade-off between the costs and effects on the consistency score in Figure~\ref{fig:tradeoff}. 
We also partially remove the zero paddings from different numbers of layers to picture the trade-off varies more accurately, where the padding type is called ``Padding-Less''. 

As shown in Figure~\ref{fig:tradeoff}, we find that the more paddings are removed, the more time agnostic the encoder becomes. 
Note that when removing all the paddings and concatenating enough real frames to the input, we obtain a perfectly time-agnostic encoder that achieves the equivariance score $=1$. 
However, it also significantly increases the memory and computational costs of the training. 
As for the other padding types, although the reflect and circular paddings provide more realistic video changes, they could drift further from the real frames and yield a smaller equivariance score than the replicate padding. 
For example, a walking person is more likely to stop than walk backward in the following frames. We find that the replicate padding, which gives $0.75$ consistency score, already resolves the time-dependence issue well in practice. Given the little extra cost it brings, we use replicate paddings and no real frames in our experiments. 

\subsection{Details of the interpolation attention}
\label{sec:attention}

\begin{figure}
    \centering
    \includegraphics[width=\linewidth]{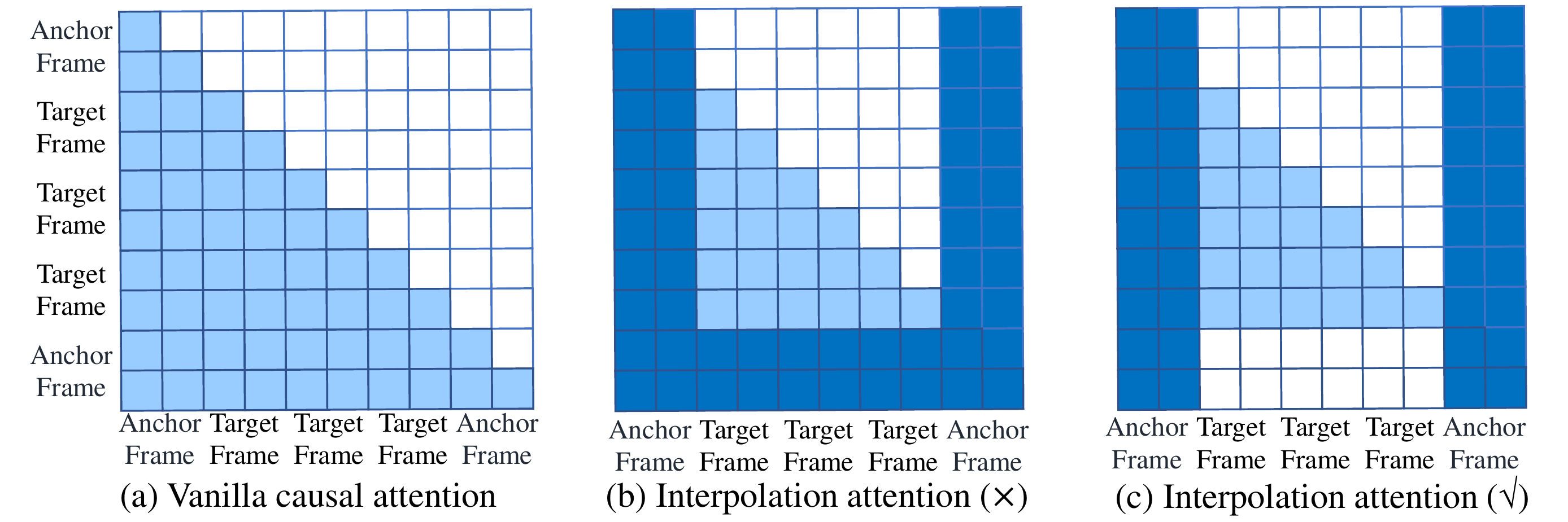}
    \caption{Illustration of the vanilla causal attention and the interpolation causal attention. For simplification, every frame is assumed to have 2 tokens.}
    \label{fig:attention}
\end{figure}

We implement the interpolation transformer based on designed attention called interpolation causal attention, as shown in the right scheme of Figure~\ref{fig:attention}(c). 
Specifically, the anchor frame (dark) is given during the inference, and the target frame (light) needs to be generated. 
In the vanilla causal attention shown in Figure~\ref{fig:attention}(a), tokens attend to the tokens \emph{in front of} it. 
In the interpolation causal attention, tokens attend to \emph{both} the tokens before it and the anchor tokens, which allows acquiring information from the anchor frames at both ends generated by the autoregressive transformer. 
We want to stress that it is important to not attend the last anchor frame on the frames to be generated like the one in the Figure~\ref{fig:attention}(b), since a multi-layer self-attention will form a shortcut and leak the information of the frames in the middle to themselves and cause the training to collapse.

\section{Experiment setups}
In this section, we provide additional details of our experiments.
\subsection{Dataset and evaluation details}
\begin{figure}[t]
    \centering
    \includegraphics[width=\linewidth]{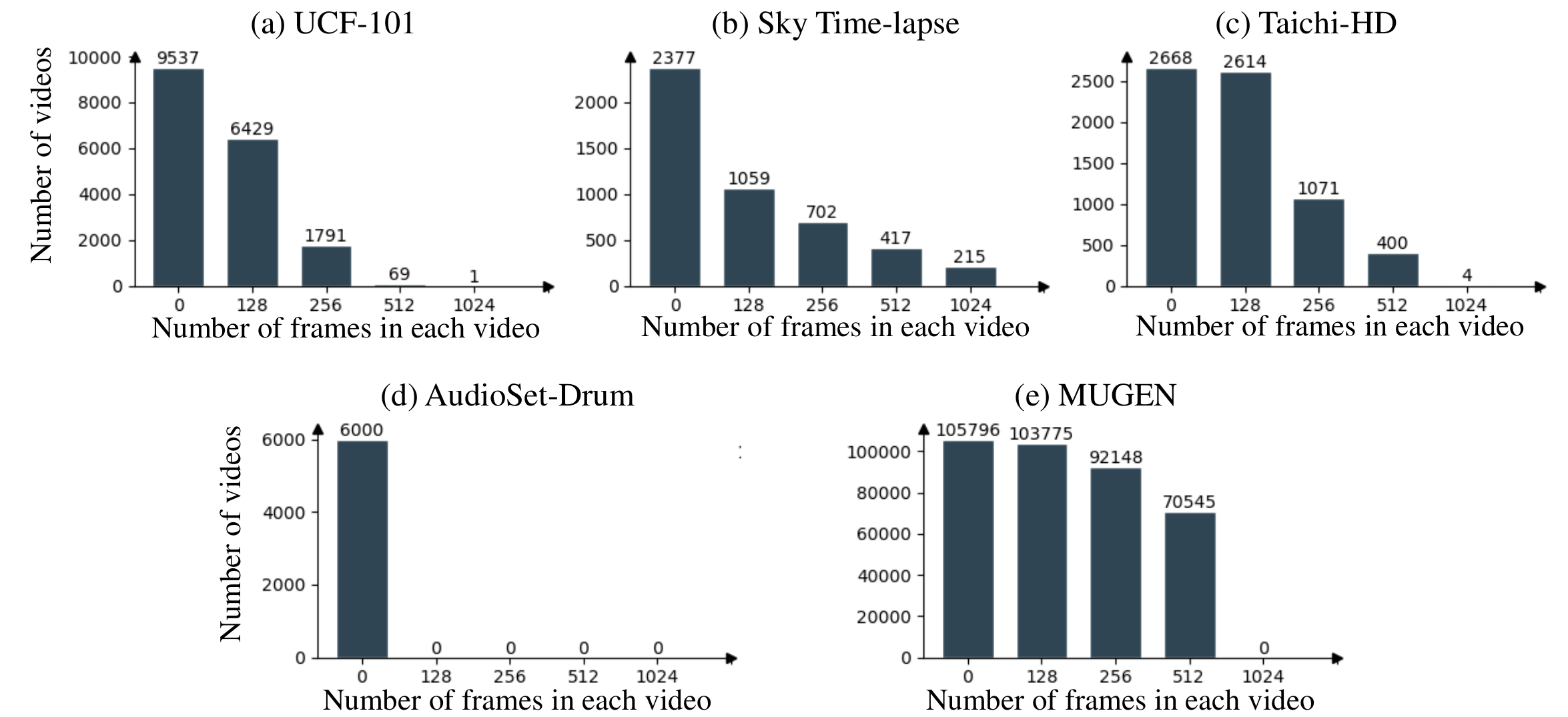}
    \caption{\textbf{Dataset statistics.} Distribution of the number of videos in each dataset that have at least a certain number of frames.}
    \label{tab:data-stats}
\end{figure}

\begin{figure}[t]
    \centering
    \includegraphics[trim=120 0 120 50,clip,width=\linewidth]{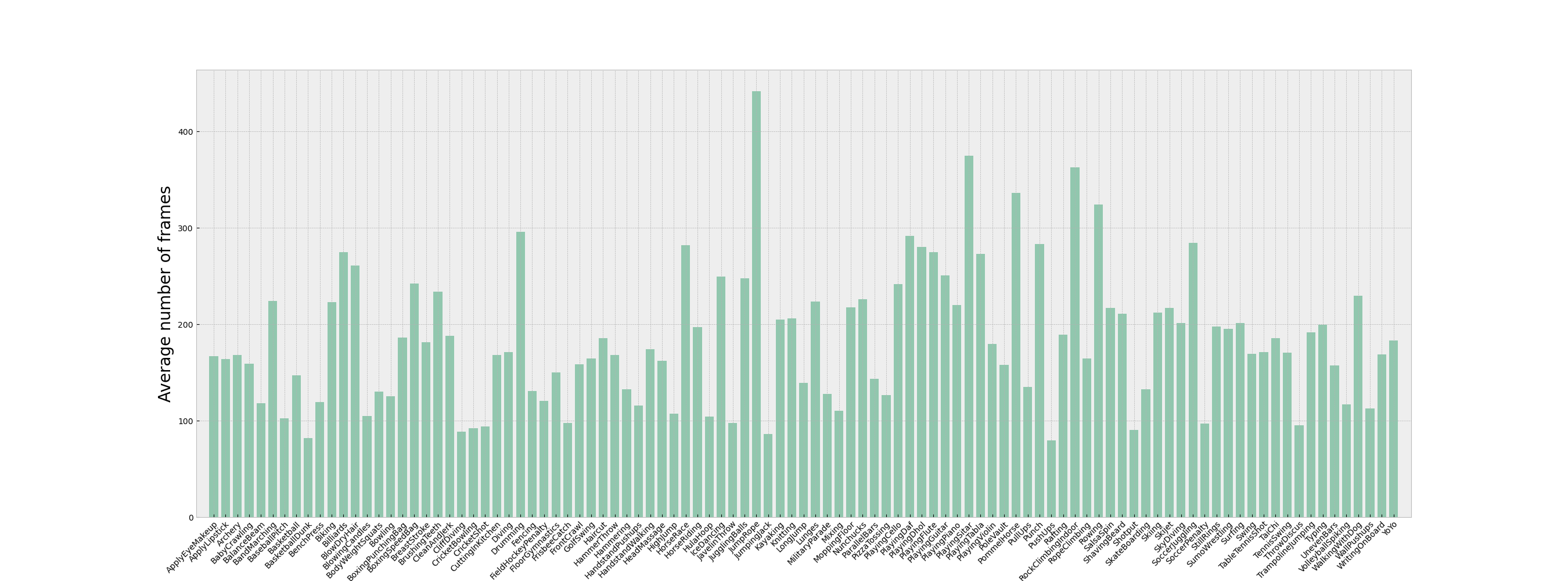}
    \caption{Average number of frames for each of class in the UCF101 dataset.}
    \label{fig:ucf_stats}
\end{figure}

We validate our approach on the UCF-101~\cite{soomro2012ucf101}, Sky Time-lapse~\cite{xiong2018learning}, Taichi-HD~\cite{siarohin2019first}, AudioSet-Drum~\cite{gemmeke2017audio}, and MUGEN~\cite{mugen2022mugen} datasets. 
Below we provide basic descriptions of these datasets and our data processing steps on each of them. 
We also report the relevant dataset statistics such as the number of long videos and the number of frames per video in Figure~\ref{tab:data-stats}.
\label{sec:dataset_appendix}
\begin{itemize}
    \item \textbf{UCF-101} is a dataset designed for action recognition which contains 101 classes and $13,320$ videos in total. 
    We train our model on its \emph{train split} which contains $9,537$ videos following the official splits.\footnote{\url{https://www.crcv.ucf.edu/data/UCF101.php}} 
    We also report number of frames in videos of every class in Figure~\ref{fig:ucf_stats}.
    \item \textbf{Sky Time-lapse} contains time-lapse videos that depict the sky under different time and weather conditions. 
    The paper~\cite{xiong2018learning} claims that $5,000$ videos are collected but only $2,647$ are actually released in the official dataset.\footnote{\url{https://github.com/weixiong-ur/mdgan}} 
    We follow \cite{yu2021generating} to train our model on the train split and test using videos from the test split.
    \item \textbf{Taichi-HD} has $2,668$ videos in total recording a single person performing Taichi.\footnote{\url{https://github.com/AliaksandrSiarohin/first-order-model/blob/master/data/taichi-loading/README.md}} 
    We follow \cite{yu2021generating} to sample frames from every $4$ frames when training our TATS-base model for a fair comparison. 
    However, training TATS-Hierarchical with this setting would drop $60\%$ of videos for not having enough frames as shown in Figure~\ref{tab:data-stats}. 
    Therefore we do not skip frames for TATS-Hierarchical.
    \item \textbf{AudioSet-Drum} is a collection of drum kit videos with audio available in the dataset. 
    The train split contains $6,000$ videos, and the test split contains $1,000$ videos. 
    All the video clips have $90$ frames. 
    We use the STFT features extracted by \cite{chatterjee2020sound2sight} as the audio data\footnote{\url{https://sites.google.com/site/metrosmiles/research/research-projects/sound2sight}}, and follow its evaluation setting to measure the image quality of the $45^\text{th}$ frames on the test set.
    \item \textbf{MUGEN} is a dataset containing videos collected from the open-sourced platform game CoinRun~\cite{cobbe2019quantifying} by recording the gameplay of trained RL agents. 
    We use their template-based algorithm auto-text, which generates textual descriptions for videos with arbitrary lengths. 
The train and test splits contain $104,796$ and $11,802$ videos, respectively, each at 3.2s to 21s (96 to 602 frames) long.
\end{itemize}
To calculate the VFDs and DVDs, we generate $2,048$ and $512$ videos for short and long video evaluation, respectively, considering time cost. 
To calculate the IS, we generate $10,000$ videos. 
To calculate the CCS and ICS for long video evaluation, we also generate $512$ videos. 
We run the evaluations for $10$ times and report the standard deviations.
\subsection{Training and inference details}
\topic{VQGAN.} 
Suggested by the VQGAN training recipe~\cite{esser2021taming}, we start GAN losses after the reconstruction loss generally converges after $10K$ steps. 
We adopt a codebook with vocabulary size  $K=16,384$ and embedding size $c=256$. 
We do not use the random start trick for codebook embeddings~\cite{dhariwal2020jukebox,yan2021videogpt}. 
Following \cite{wang2018video}, we set $\lambda_\text{rec}=\lambda_\text{match}=4.0$ and $\lambda_\text{disc}=1.0$. 
We use the ADAM optimizer~\cite{kingma2014adam} with $lr = 3e^{-5}$ and $(\beta_1, \beta_2) = (0.5, 0.9)$. 
As discussed in Section~\ref{sec:video_vqgan}, we always clip the gradients to have euclidean norm $=1$. 
We train the VQGAN on $8$ NVIDIA V100 32GB GPUs with batch size $=2$ on each gpu and accumulated batches $=6$ for $30K$ steps. 
Each model usually takes around $57$ hours to train.

\topic{Transformers.} 
Both autoregressive and interpolation transformers contain $24$ layers, $16$ heads, and embedding size $=1024$. 
We use the AdamW optimizer~\cite{loshchilov2018decoupled} with a base learning rate of $4.5e^{-6}$, where we linearly scale up by the total batch size. 
We train the transformers on $8$ NVIDIA V100 32GB GPUs with batch size $=3$ on each GPU until the training loss saturates. 
Specifically, we train the autoregressive transformers for $500K$ steps as a general setting except that we train TATS-base on the UCF-101 dataset for $1.35M$ as it keeps improving the results. It takes around $10$ days to train the transformer models.
In practice, we find that the interpolation setting simplifies the problem dramatically, so we only train the interpolation transformers for $30K$ for the best generalization performance. 
At the inference time, we adopt sampling strategy where temperature $t=1$, top-$k$ with $k=2048$, and top-$p$ with $p=0.80$ as our general setting (for interpolation transformers, a smaller $q$ and $k$ usually produces better results). Sampling $1$ video with $1024$ frames takes around $30$ minutes using TATS-base on a single Quadro P6000 GPU, while TATS-hierarchical reduces this time to $7.5$ minutes for autoregressive transformer and $23$ seconds for interpolation transformer.

\subsubsection{Baselines}
Apart from those that have been discussed in the related work section, we provide additional descriptions on the baselines we compared with and other GAN-based video generation models in this section. TGAN~\cite{saito2017temporal} proposes to generate a fixed number of latent vectors as input to an image generator to synthesize the corresponding frames. MoCoGAN~\cite{Tulyakov_2018_CVPR} utilizes a RNN to sample motion vectors to synthesize different frames. MoCoGAN-HD~\cite{tian2021a} leverages a LSTM to predict a trajectory in the latent space of a trained image generator.  DVD-GAN~\cite{clark2019adversarial} adopts a similar architecture as MoCoGAN with a focus on scaling up training. TSB~\cite{munoz2021temporal} further improves  MoCoGAN by mixing information of adjacent frames using an operation called temporal shift. 
TGAN2~\cite{saito2020train} proposes to divide the generator into multiple small sub-generators and introduces a subsampling layer that reduces the frame rate between each pair of consecutive sub-generators. 
HVG~\cite{castrejon2021hierarchical} further introduces a hierarchical pipeline to interpolate and upsample low-resolution and low-frame-rate videos gradually. 
ProgressiveVGAN~\cite{acharya2018towards} extends ProgressiveGAN~\cite{karras2018progressive} to video generation by simultaneously generating in the temporal direction progressively. LDVD-GAN~\cite{kahembwe2020lower} analyzes the discriminators used in video GANs and designs improved discriminators considering the convolution kernel dimensionality. CCVS~\cite{le2021ccvs} utilizes VQVAE to compress frames for training transformer and a flow module to improve temporal consistency. VideoGPT~\cite{yan2021videogpt} further leverages 3D convolution and axial self-attention in the VQVAE to encode the temporal dimension as well. 
DIGAN~\cite{yu2021generating} first generalized the idea of implicit neural representations to video generation by decomposing the network weights for the spatial and temporal coordinates. StyleGAN-v~\cite{skorokhodov2021stylegan} maps a continuous temporal positional embedding to the input feature map of the StyleGAN.

\subsection{Comparison of the computational costs}

\cameraready{
We compute the time to generate a single video of $1024$ frames using different methods in Table~\ref{tab:inference_time}. 
The slow inference speed of autoregressive models is often criticized. 
However, among all the methods that build on VQVAE and transformer framework, our TATS-hierarchical is the fastest -- $1/3$ the time of CCVS and $1/5$ the time of VideoGPT. 
Accelerating autoregressive transformers is an active research area. 
Further improvements can be achieved by methods such as sparse attention, which we leave for future exploration.
}

\begin{table}[h]
\centering
\caption{Time for generating a 1024-frame video.}
\addtolength{\tabcolsep}{3.5pt}
\label{tab:inference_time}
\begin{tabular}{llllll}
\toprule
DIGAN & MoCoGAN-HD & CCVS & VideoGPT & TATS-base & TATS-hierarchical \\
\midrule
4.2 sec  & 28.5 sec  &22 min & 42 min & 30 min & 7.5 min \\
\bottomrule
\end{tabular}
\end{table}

\section{Additional results on long video generation}

\subsection{More videos with repeated actions and smooth transitions.}
Video generation results with repeated actions can be widely seen in other UCF-101 classes as well as shown in Figure~\ref{fig:more_intersection}. 
In addition to the generated sky videos with smooth transitions, we also show that there are such cases in UCF-101 and Taichi-HD datasets in Figure~\ref{fig:intersection_taichi_ucf}. 
We can see in the example of UCF-101 \textit{Sky Diving} video that the transition proceeds clearly with unrealistic content as only limited data are available per class. 
In the example of Taichi video, we can see that the color of the pants transforms smoothly. 
However, there is still unrealistic content in such videos. Therefore, we argue that a split of content and motion generation could be helpful in these cases~\cite {Tulyakov_2018_CVPR,yu2021generating,tian2021a}.

\subsection{Failure cases.} 
Errors could occur and accumulate during the application of sliding window attention. 
We show several typical failure examples in Figure~\ref{fig:failure_case}. 
In the example of  \textit{Boxing Punching Bag}, an error occurs at around $300$ frames, and the general quality of the video deteriorates after that. 
Repeated tokens are generated in the deteriorated area, which is a commonly observed issue in sequence generation~\cite{Holtzman2020The,jiang2018sequence}. 
We also find that this kind of issue happens more often in the areas which contain large motions. 
The video quality could also degrade quickly if the thematic event of the video has a clear end. For example, the generated \textit{Long Jump} video quickly transits to the other scene and degenerates when the person finishes the action.

\begin{figure}[h]
    \centering
    \includegraphics[width=0.85\linewidth]{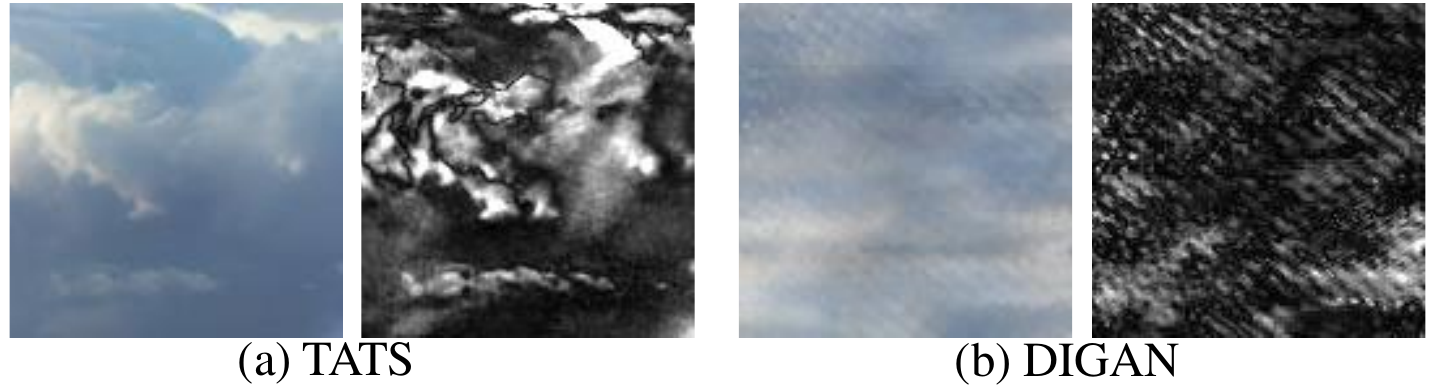}
    \caption{The $1000^\text{th}$ generated frame and the average pixel difference w.r.t. the $1024^\text{th}$ frame. DIGAN generations contain clear periodic artifacts.}
    \label{fig:displacement}
\end{figure}

\subsection{Additional comparison with DIGAN.} \cameraready{
We find that DIGAN presents clear quality degradation on the Sky dataset. To show this, we calculate the average difference between the $1000^\text{th}$ and $1024^\text{th}$ frames generated by DIGAN and our method in Figure~\ref{fig:displacement}. 
DIGAN results show periodic artifacts induced by the sinusoidal positional encodings (repeated changes in the diagonal direction). 
However, this is not reflected in the FVD metric, which we conjecture is due to the domain gap with respect to the Kinetics dataset which the I3D model is trained on. 
So to quantify this, we conduct human evaluation to compare TATS with DIGAN. 
We randomly sampled 100 generated videos for each method
and cropped the videos to the last 128 frames. 
We showed raters a pair of videos (one from each method), and asked them to select the video with fewer artifacts. 
Each video pair was evaluated by around 5 raters. 
TATS was chosen over DIGAN $67\%$ of the time ($355$ vs. $175$). 
Under a binomial test, TATS is statistically significantly better than DIGAN with confidence $>0.95$.
}


\section{Additional related works}
\topic{The implicit bias of zero padding.} 
The positional inductive bias introduced by zero padding has drawn increasing attention recently~\cite{islam2019much,kayhan2020translation,xu2021positional,alsallakh2021mind}. In most cases, such inductive bias is helpful in classification~\cite{islam2019much}, generation~\cite{xu2021positional}, or object detection~\cite{alsallakh2021mind}. Our paper observes a case where this inductive bias is harmful. 

\topic{Video prediction.} Video prediction aims at modeling the transformation between frames and predicting future frames given real frames~\cite{ranzato2014video,srivastava2015unsupervised,finn2016unsupervised,van2017transformation,kumar2019videoflow,park2021vid}. For instance, \cite{van2017transformation} divides the frames into patches, calculates the affine transformation between temporally adjacent patches, and predicts such affine transformation to be applied to the most recent frame. \cite{luc2017predicting} predicts videos of semantic maps and argues that autoregressive models lead to error propagation when more frames are predicted while being more accurate in the semantic segmentation space. Flow-based models~\cite{kumar2019videoflow} and ODE-based models~\cite{park2021vid} have also been used for video prediction. Different from video prediction, video generation focus on producing videos from noise. When applying to long videos, one substantial difference is that video prediction starts from a real frame while video generation starts from a generated frame. Some video generation models fall into the middle part~\cite{le2021ccvs} that predicts future frames given frames generated by an image generator. We have shown that these models also suffer from quality degradation.

\topic{Conditional video generation.} 
Text-conditioned video generation has been studied in multiple papers. SyncDraw~\cite{mittal2017sync} first proposes to combine VAE and RNN for video generation while conditioning on texts. T2V~\cite{li2018video} uses CVAE to generate the gist then GANs with 3D convolutions to generate fixed-length low-resolution videos. The text is encoded in a convolutions filter to process the gist. TFGAN~\cite{balaji2019conditional} proposes a multi-scale text-conditioning discriminator and follows MoCoGAN~\cite{Tulyakov_2018_CVPR} to use an RNN to model the temporal information. IRC-GAN~\cite{deng2019irc} proposes to use a recurrent transconvolutional generator and mutual-information introspection to generate videos based text. Craft~\cite{Gupta2018ImagineTS} sequentially composes a scene layout and retrieves entities from a video database to create complex scene videos. GODIVA~\cite{wu2021godiva} relies on a pretrained VQVAE model to produce video frame representations and autoregressively predicts the video representations based on the input text representations and former frames. Each element's prediction in the current frame's representation attends to the previous row, column, and time. As for audio-conditioned video generation, Sound2Sight~\cite{chatterjee2020sound2sight} proposes to model the previous frames and audios with a multi-head audio-visual transformer and predicts future frames with a prediction network based on sequence-to-sequence architecture. Vougioukas et al.~\cite{vougioukas2018end} studies speech-driven facial animation, which aims at generating face videos given the speech audios. They propose a framework based on RNNs and a frame generator. Some conditional generation frameworks are also able to generate long videos either when minimal changes occur in the video scenes~\cite{menapace2021playable} or strong conditional information is given such as segmentation maps~\cite{wang2018video,mallya2020world}. Our paper focuses on more realistic and complex videos with weak or no conditional information available. 

\begin{figure}[h]
    \centering
    \vspace{-.5cm}
    \includegraphics[trim=0 0 0 0,clip,width=\linewidth]{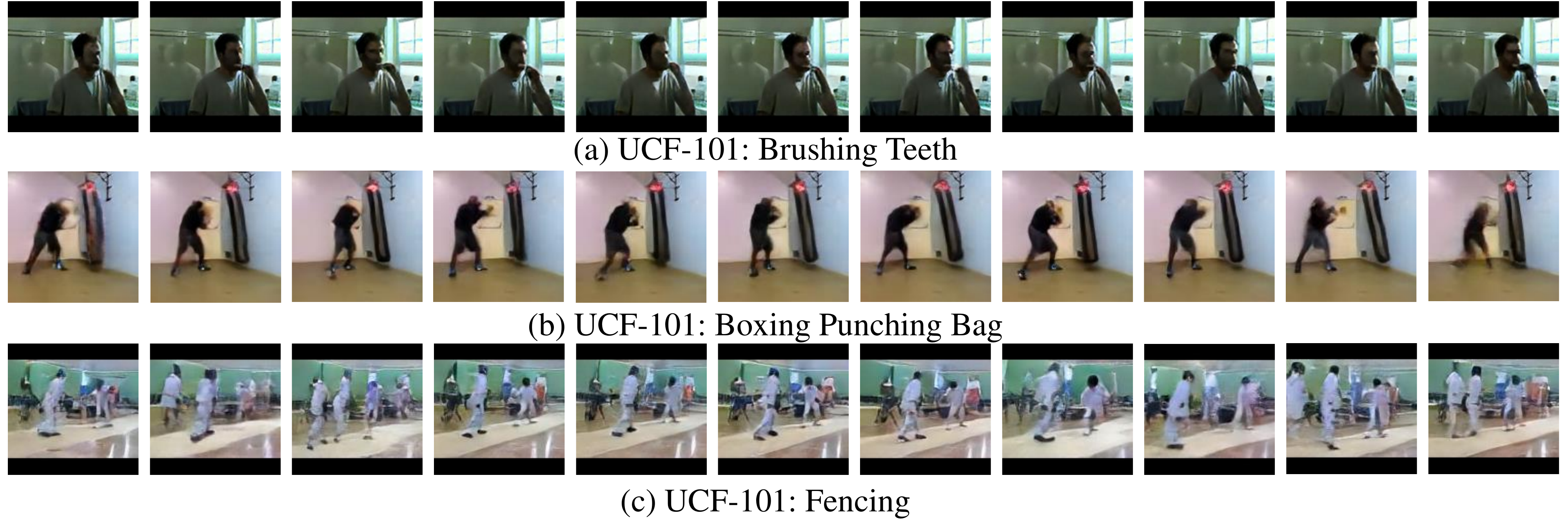}
    \caption{More class-conditional generation results of UCF-101 videos with $1,024$ frames that contains repeated action.}
    \label{fig:more_intersection}
\end{figure}

\begin{figure}[h]
    \centering
    \vspace{-.5cm}
    \includegraphics[trim=0 0 0 0,clip,width=\linewidth]{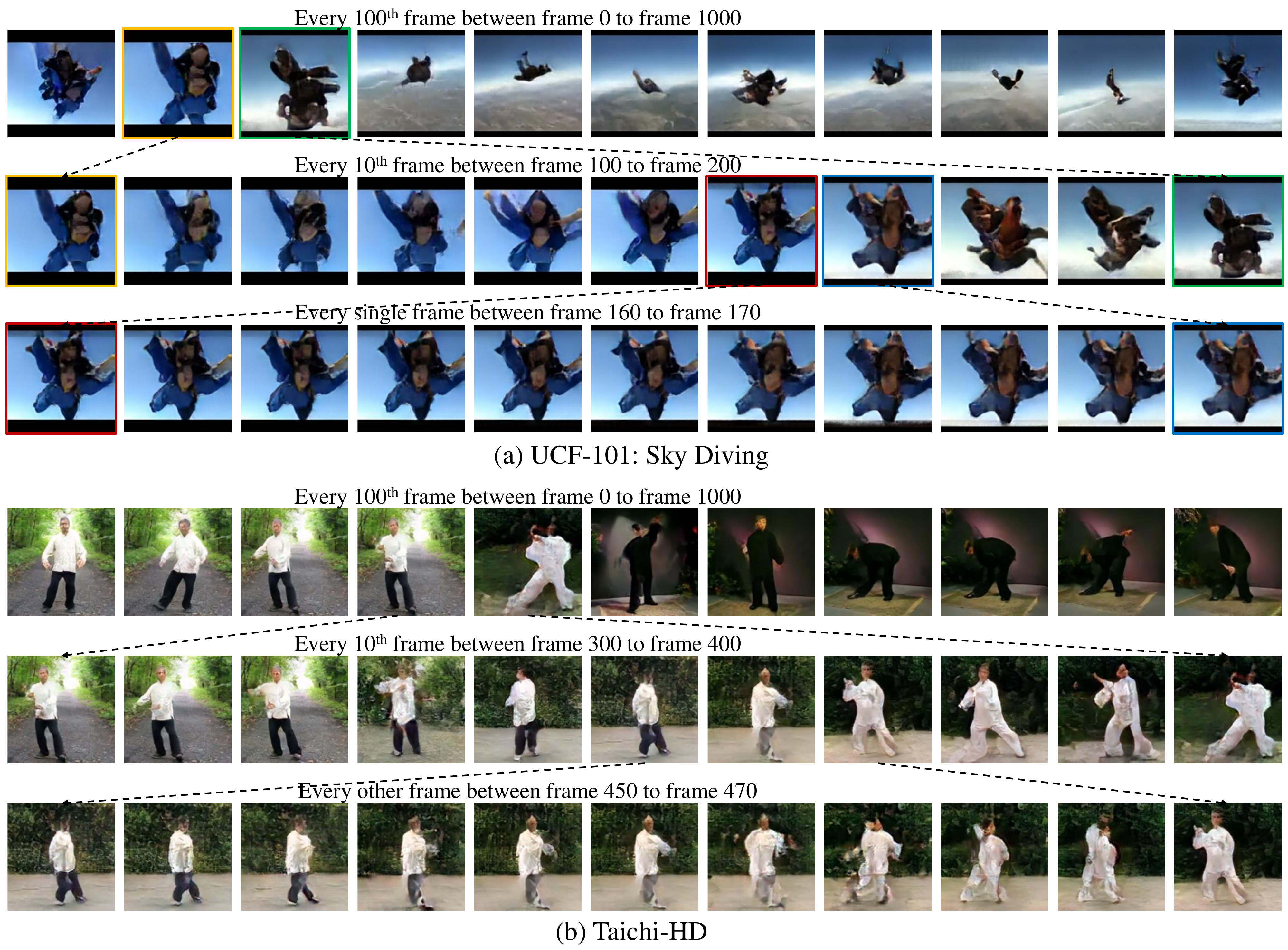}
    \caption{Unconditional and class-conditional generation results of \textit{Sky Diving} and Taichi-HD videos with $1,024$ frames that contain smooth transitions.}
    \label{fig:intersection_taichi_ucf}
\end{figure}

\begin{figure}[h]
    \centering
    \vspace{-.5cm}
    \includegraphics[width=\linewidth]{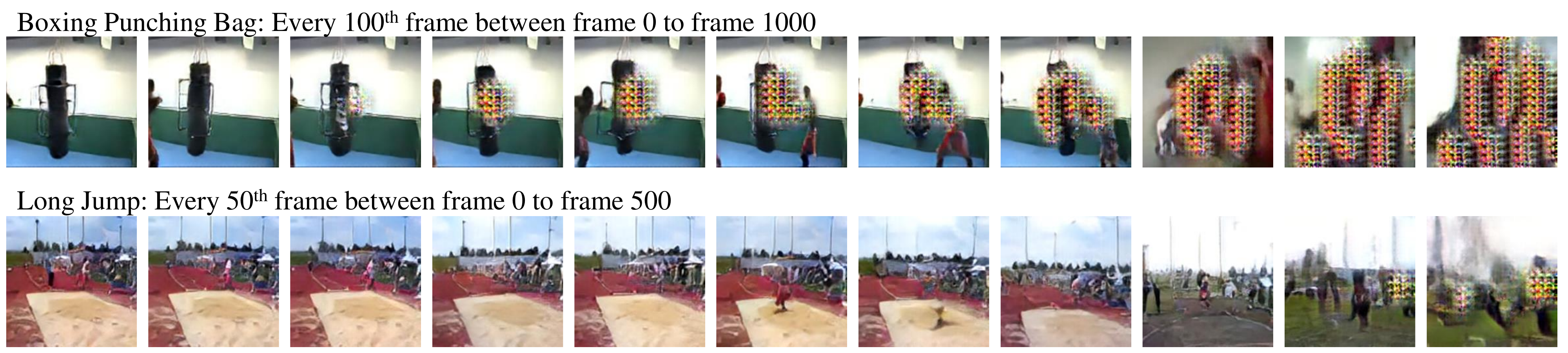}
    \caption{Failure cases of class-conditional generation results of long videos on the UCF-101 dataset.}
    \label{fig:failure_case}
\end{figure}


\clearpage
%
%
\bibliographystyle{splncs04}
\bibliography{egbib}

\begin{thebibliography}{10}
\providecommand{\url}[1]{\texttt{#1}}
\providecommand{\urlprefix}{URL }
\providecommand{\doi}[1]{https://doi.org/#1}

\bibitem{acharya2018towards}
Acharya, D., Huang, Z., Paudel, D.P., Van~Gool, L.: Towards high resolution
  video generation with progressive growing of sliced wasserstein gans. arXiv
  preprint arXiv:1810.02419  (2018)

\bibitem{alsallakh2021mind}
Alsallakh, B., Kokhlikyan, N., Miglani, V., Yuan, J., Reblitz-Richardson, O.:
  Mind the pad -- {CNN}s can develop blind spots. In: ICLR (2021)

\bibitem{balaji2019conditional}
Balaji, Y., Min, M.R., Bai, B., Chellappa, R., Graf, H.P.: Conditional gan with
  discriminative filter generation for text-to-video synthesis. In: IJCAI
  (2019)

\bibitem{brock2018large}
Brock, A., Donahue, J., Simonyan, K.: Large scale gan training for high
  fidelity natural image synthesis. In: ICLR (2018)

\bibitem{brooks2022generating}
Brooks, T., Hellsten, J., Aittala, M., Wang, T.C., Aila, T., Lehtinen, J., Liu,
  M.Y., Efros, A.A., Karras, T.: Generating long videos of dynamic scenes.
  arXiv preprint arXiv:2206.03429  (2022)

\bibitem{brown2020language}
Brown, T., Mann, B., Ryder, N., Subbiah, M., Kaplan, J.D., Dhariwal, P.,
  Neelakantan, A., Shyam, P., Sastry, G., Askell, A., et~al.: Language models
  are few-shot learners. NeurIPS  (2020)

\bibitem{carreira2017quo}
Carreira, J., Zisserman, A.: Quo vadis, action recognition? a new model and the
  kinetics dataset. In: CVPR (2017)

\bibitem{castrejon2021hierarchical}
Castrejon, L., Ballas, N., Courville, A.: Hierarchical video generation for
  complex data. arXiv preprint arXiv:2106.02719  (2021)

\bibitem{chatterjee2020sound2sight}
Chatterjee, M., Cherian, A.: Sound2sight: Generating visual dynamics from sound
  and context. In: ECCV (2020)

\bibitem{child2020very}
Child, R.: Very deep vaes generalize autoregressive models and can outperform
  them on images. In: ICLR (2020)

\bibitem{clark2019adversarial}
Clark, A., Donahue, J., Simonyan, K.: Adversarial video generation on complex
  datasets. arXiv preprint arXiv:1907.06571  (2019)

\bibitem{cobbe2019quantifying}
Cobbe, K., Klimov, O., Hesse, C., Kim, T., Schulman, J.: Quantifying
  generalization in reinforcement learning. In: ICML (2019)

\bibitem{deng2019irc}
Deng, K., Fei, T., Huang, X., Peng, Y.: Irc-gan: Introspective recurrent
  convolutional gan for text-to-video generation. In: IJCAI (2019)

\bibitem{dhariwal2020jukebox}
Dhariwal, P., Jun, H., Payne, C., Kim, J.W., Radford, A., Sutskever, I.:
  Jukebox: A generative model for music. arXiv preprint arXiv:2005.00341
  (2020)

\bibitem{esser2021taming}
Esser, P., Rombach, R., Ommer, B.: Taming transformers for high-resolution
  image synthesis. In: CVPR (2021)

\bibitem{fan2018hierarchical}
Fan, A., Lewis, M., Dauphin, Y.: Hierarchical neural story generation (2018)

\bibitem{finn2016unsupervised}
Finn, C., Goodfellow, I., Levine, S.: Unsupervised learning for physical
  interaction through video prediction. NeurIPS  (2016)

\bibitem{gemmeke2017audio}
Gemmeke, J.F., Ellis, D.P., Freedman, D., Jansen, A., Lawrence, W., Moore,
  R.C., Plakal, M., Ritter, M.: Audio set: An ontology and human-labeled
  dataset for audio events. In: ICASSP (2017)

\bibitem{goodfellow2014generative}
Goodfellow, I., Pouget-Abadie, J., Mirza, M., Xu, B., Warde-Farley, D., Ozair,
  S., Courville, A., Bengio, Y.: Generative adversarial nets. NeurIPS  (2014)

\bibitem{Gupta2018ImagineTS}
Gupta, T., Schwenk, D., Farhadi, A., Hoiem, D., Kembhavi, A.: Imagine this!
  scripts to compositions to videos. In: ECCV (2018)

\bibitem{mugen2022mugen}
Hayes, T., Zhang, S., Yin, X., Pang, G., Sheng, S., Yang, H., Ge, S., Hu, Q.,
  Parikh, D.: Mugen: A playground for video-audio-text multimodal understanding
  and generation. arXiv preprint arXiv:2204.08058  (2022)

\bibitem{ho2022video}
Ho, J., Salimans, T., Gritsenko, A., Chan, W., Norouzi, M., Fleet, D.J.: Video
  diffusion models. arXiv preprint arXiv:2204.03458  (2022)

\bibitem{Holtzman2020The}
Holtzman, A., Buys, J., Du, L., Forbes, M., Choi, Y.: The curious case of
  neural text degeneration. In: ICLR (2020)

\bibitem{hong2022cogvideo}
Hong, W., Ding, M., Zheng, W., Liu, X., Tang, J.: Cogvideo: Large-scale
  pretraining for text-to-video generation via transformers. arXiv preprint
  arXiv:2205.15868  (2022)

\bibitem{islam2019much}
Islam, M.A., Jia, S., Bruce, N.D.: How much position information do
  convolutional neural networks encode? In: ICLR (2019)

\bibitem{jiang2018sequence}
Jiang, S., de~Rijke, M.: Why are sequence-to-sequence models so dull?
  understanding the low-diversity problem of chatbots. In: EMNLP Workshop
  (2018)

\bibitem{johnson2016perceptual}
Johnson, J., Alahi, A., Fei-Fei, L.: Perceptual losses for real-time style
  transfer and super-resolution. In: ECCV (2016)

\bibitem{kahembwe2020lower}
Kahembwe, E., Ramamoorthy, S.: Lower dimensional kernels for video
  discriminators. Neural Networks  (2020)

\bibitem{kalchbrenner2017video}
Kalchbrenner, N., Oord, A., Simonyan, K., Danihelka, I., Vinyals, O., Graves,
  A., Kavukcuoglu, K.: Video pixel networks. In: ICML (2017)

\bibitem{karras2018progressive}
Karras, T., Aila, T., Laine, S., Lehtinen, J.: Progressive growing of gans for
  improved quality, stability, and variation. In: ICLR (2018)

\bibitem{karras2021alias}
Karras, T., Aittala, M., Laine, S., H{\"a}rk{\"o}nen, E., Hellsten, J.,
  Lehtinen, J., Aila, T.: Alias-free generative adversarial networks. NeurIPS
  (2021)

\bibitem{karras2019style}
Karras, T., Laine, S., Aila, T.: A style-based generator architecture for
  generative adversarial networks. In: CVPR (2019)

\bibitem{karras2020analyzing}
Karras, T., Laine, S., Aittala, M., Hellsten, J., Lehtinen, J., Aila, T.:
  Analyzing and improving the image quality of stylegan. In: CVPR (2020)

\bibitem{kayhan2020translation}
Kayhan, O.S., Gemert, J.C.v.: On translation invariance in cnns: Convolutional
  layers can exploit absolute spatial location. In: CVPR (2020)

\bibitem{kingma2014adam}
Kingma, D.P., Ba, J.: Adam: A method for stochastic optimization. ICLR  (2015)

\bibitem{kumar2019videoflow}
Kumar, M., Babaeizadeh, M., Erhan, D., Finn, C., Levine, S., Dinh, L., Kingma,
  D.: Videoflow: A conditional flow-based model for stochastic video
  generation. In: ICLR (2019)

\bibitem{le2021ccvs}
Le~Moing, G., Ponce, J., Schmid, C.: Ccvs: Context-aware controllable video
  synthesis. NeurIPS  (2021)

\bibitem{lee2021vitgan}
Lee, K., Chang, H., Jiang, L., Zhang, H., Tu, Z., Liu, C.: Vitgan: Training
  gans with vision transformers. arXiv preprint arXiv:2107.04589  (2021)

\bibitem{li2018video}
Li, Y., Min, M., Shen, D., Carlson, D., Carin, L.: Video generation from text.
  In: AAAI (2018)

\bibitem{loshchilov2018decoupled}
Loshchilov, I., Hutter, F.: Decoupled weight decay regularization. In:
  International Conference on Learning Representations (2018)

\bibitem{luc2020transformation}
Luc, P., Clark, A., Dieleman, S., Casas, D.d.L., Doron, Y., Cassirer, A.,
  Simonyan, K.: Transformation-based adversarial video prediction on
  large-scale data. arXiv preprint arXiv:2003.04035  (2020)

\bibitem{luc2017predicting}
Luc, P., Neverova, N., Couprie, C., Verbeek, J., LeCun, Y.: Predicting deeper
  into the future of semantic segmentation. In: ICCV (2017)

\bibitem{mallya2020world}
Mallya, A., Wang, T.C., Sapra, K., Liu, M.Y.: World-consistent video-to-video
  synthesis. In: ECCV (2020)

\bibitem{menapace2021playable}
Menapace, W., Lathuili{\`e}re, S., Tulyakov, S., Siarohin, A., Ricci, E.:
  Playable video generation. In: CVPR (2021)

\bibitem{mittal2017sync}
Mittal, G., Marwah, T., Balasubramanian, V.N.: Sync-draw: Automatic video
  generation using deep recurrent attentive architectures. In: MM (2017)

\bibitem{munoz2021temporal}
Munoz, A., Zolfaghari, M., Argus, M., Brox, T.: Temporal shift gan for large
  scale video generation. In: WACV (2021)

\bibitem{nash2022transframer}
Nash, C., Carreira, J., Walker, J., Barr, I., Jaegle, A., Malinowski, M.,
  Battaglia, P.: Transframer: Arbitrary frame prediction with generative
  models. arXiv preprint arXiv:2203.09494  (2022)

\bibitem{van2017neural}
van~den Oord, A., Vinyals, O., Kavukcuoglu, K.: Neural discrete representation
  learning. In: NeurIPS (2017)

\bibitem{park2021vid}
Park, S., Kim, K., Lee, J., Choo, J., Lee, J., Kim, S., Choi, Y.: Vid-ode:
  Continuous-time video generation with neural ordinary differential equation.
  In: AAAI. AAAI (2021)

\bibitem{radford2021learning}
Radford, A., Kim, J.W., Hallacy, C., Ramesh, A., Goh, G., Agarwal, S., Sastry,
  G., Askell, A., Mishkin, P., Clark, J., et~al.: Learning transferable visual
  models from natural language supervision. In: ICML (2021)

\bibitem{radford2018improving}
Radford, A., Narasimhan, K., Salimans, T., Sutskever, I.: Improving language
  understanding by generative pre-training

\bibitem{radford2019language}
Radford, A., Wu, J., Child, R., Luan, D., Amodei, D., Sutskever, I.: Language
  models are unsupervised multitask learners  (2019)

\bibitem{rakhimov2020latent}
Rakhimov, R., Volkhonskiy, D., Artemov, A., Zorin, D., Burnaev, E.: Latent
  video transformer. arXiv preprint arXiv:2006.10704  (2020)

\bibitem{ramesh2021zero}
Ramesh, A., Pavlov, M., Goh, G., Gray, S., Voss, C., Radford, A., Chen, M.,
  Sutskever, I.: Zero-shot text-to-image generation. arXiv preprint
  arXiv:2102.12092  (2021)

\bibitem{ranzato2014video}
Ranzato, M., Szlam, A., Bruna, J., Mathieu, M., Collobert, R., Chopra, S.:
  Video (language) modeling: a baseline for generative models of natural
  videos. arXiv preprint arXiv:1412.6604  (2014)

\bibitem{saito2017temporal}
Saito, M., Matsumoto, E., Saito, S.: Temporal generative adversarial nets with
  singular value clipping. In: ICCV (2017)

\bibitem{saito2020train}
Saito, M., Saito, S., Koyama, M., Kobayashi, S.: Train sparsely, generate
  densely: Memory-efficient unsupervised training of high-resolution temporal
  gan. IJCV  (2020)

\bibitem{siarohin2019first}
Siarohin, A., Lathuili{\`e}re, S., Tulyakov, S., Ricci, E., Sebe, N.: First
  order motion model for image animation. NeurIPS  (2019)

\bibitem{simonyan2014very}
Simonyan, K., Zisserman, A.: Very deep convolutional networks for large-scale
  image recognition. arXiv preprint arXiv:1409.1556  (2014)

\bibitem{sitzmann2020implicit}
Sitzmann, V., Martel, J., Bergman, A., Lindell, D., Wetzstein, G.: Implicit
  neural representations with periodic activation functions. NeurIPS  (2020)

\bibitem{skorokhodov2021stylegan}
Skorokhodov, I., Tulyakov, S., Elhoseiny, M.: Stylegan-v: A continuous video
  generator with the price, image quality and perks of stylegan2. arXiv
  preprint arXiv:2112.14683  (2021)

\bibitem{soomro2012ucf101}
Soomro, K., Zamir, A.R., Shah, M.: Ucf101: A dataset of 101 human actions
  classes from videos in the wild. arXiv preprint arXiv:1212.0402  (2012)

\bibitem{srivastava2015unsupervised}
Srivastava, N., Mansimov, E., Salakhudinov, R.: Unsupervised learning of video
  representations using lstms. In: ICML (2015)

\bibitem{tancik2020fourier}
Tancik, M., Srinivasan, P., Mildenhall, B., Fridovich-Keil, S., Raghavan, N.,
  Singhal, U., Ramamoorthi, R., Barron, J., Ng, R.: Fourier features let
  networks learn high frequency functions in low dimensional domains. NeurIPS
  (2020)

\bibitem{tian2021a}
Tian, Y., Ren, J., Chai, M., Olszewski, K., Peng, X., Metaxas, D.N., Tulyakov,
  S.: A good image generator is what you need for high-resolution video
  synthesis. In: ICLR (2021)

\bibitem{tran2015learning}
Tran, D., Bourdev, L., Fergus, R., Torresani, L., Paluri, M.: Learning
  spatiotemporal features with 3d convolutional networks. In: ICCV (2015)

\bibitem{Tulyakov_2018_CVPR}
Tulyakov, S., Liu, M.Y., Yang, X., Kautz, J.: Mocogan: Decomposing motion and
  content for video generation. In: CVPR (June 2018)

\bibitem{unterthiner2018towards}
Unterthiner, T., van Steenkiste, S., Kurach, K., Marinier, R., Michalski, M.,
  Gelly, S.: Towards accurate generative models of video: A new metric \&
  challenges. ICLR  (2019)

\bibitem{van2017transformation}
Van~Amersfoort, J., Kannan, A., Ranzato, M., Szlam, A., Tran, D., Chintala, S.:
  Transformation-based models of video sequences. arXiv preprint
  arXiv:1701.08435  (2017)

\bibitem{vondrick2016generating}
Vondrick, C., Pirsiavash, H., Torralba, A.: Generating videos with scene
  dynamics. NeurIPS  (2016)

\bibitem{vougioukas2018end}
Vougioukas, K., Petridis, S., Pantic, M.: End-to-end speech-driven facial
  animation with temporal gans. BMVC  (2018)

\bibitem{wang2018video}
Wang, T.C., Liu, M.Y., Zhu, J.Y., Liu, G., Tao, A., Kautz, J., Catanzaro, B.:
  Video-to-video synthesis. In: NeurIPS (2018)

\bibitem{wang2018high}
Wang, T.C., Liu, M.Y., Zhu, J.Y., Tao, A., Kautz, J., Catanzaro, B.:
  High-resolution image synthesis and semantic manipulation with conditional
  gans. In: CVPR (2018)

\bibitem{Weissenborn2020Scaling}
Weissenborn, D., Täckström, O., Uszkoreit, J.: Scaling autoregressive video
  models. In: ICLR (2020)

\bibitem{wu2021godiva}
Wu, C., Huang, L., Zhang, Q., Li, B., Ji, L., Yang, F., Sapiro, G., Duan, N.:
  Godiva: Generating open-domain videos from natural descriptions. arXiv
  preprint arXiv:2104.14806  (2021)

\bibitem{wu2021n}
Wu, C., Liang, J., Ji, L., Yang, F., Fang, Y., Jiang, D., Duan, N.:
  N$\backslash$" uwa: Visual synthesis pre-training for neural visual world
  creation. arXiv preprint arXiv:2111.12417  (2021)

\bibitem{wu2018group}
Wu, Y., He, K.: Group normalization. In: ECCV (2018)

\bibitem{xiong2018learning}
Xiong, W., Luo, W., Ma, L., Liu, W., Luo, J.: Learning to generate time-lapse
  videos using multi-stage dynamic generative adversarial networks. In: CVPR
  (2018)

\bibitem{xu2021positional}
Xu, R., Wang, X., Chen, K., Zhou, B., Loy, C.C.: Positional encoding as spatial
  inductive bias in gans. In: CVPR (2021)

\bibitem{yan2021videogpt}
Yan, W., Zhang, Y., Abbeel, P., Srinivas, A.: Videogpt: Video generation using
  vq-vae and transformers. arXiv preprint arXiv:2104.10157  (2021)

\bibitem{yu2021generating}
Yu, S., Tack, J., Mo, S., Kim, H., Kim, J., Ha, J.W., Shin, J.: Generating
  videos with dynamics-aware implicit generative adversarial networks. In: ICLR
  (2021)

\bibitem{zhang2018unreasonable}
Zhang, R., Isola, P., Efros, A.A., Shechtman, E., Wang, O.: The unreasonable
  effectiveness of deep features as a perceptual metric. In: CVPR (2018)

\end{thebibliography}
\end{document}